\documentclass [10pt , a4paper]{article}
\usepackage{times}

\usepackage[margin=3cm]{geometry}
\usepackage[small]{caption}
\usepackage[ansinew]{inputenc}
\usepackage{amsmath}
\usepackage{bm}
\usepackage{bbm}
\usepackage{verbatim}
\usepackage{nccmath}
\usepackage[psamsfonts]{amssymb}
\usepackage{graphicx}
\usepackage{subfigure}
\usepackage{cancel}
\usepackage{verbatim}

\newenvironment{matlab}{\comment}{\endcomment}

\title{Variational Gaussian Process Dynamical Systems}

 \author{
 Andreas Damianou \\
 Sheffield Institute for Translational Neuroscience\\
 University of Sheffield \\
\texttt{andreas.damianou@sheffield.ac.uk}
            \and
 Michalis K. Titsias \\
 School of Computer Science\\
 University of Manchester, UK \\
 \texttt{mtitsias@cs.man.ac.uk} \\
            \and
 Neil Lawrence \\
Sheffield Institute for Translational Neuroscience\\
and\\ Department of Computer Science\\
University of Sheffield\\
\texttt{neil@dcs.shef.ac.uk} 
 }

\begin{document}

\newcommand{\highlight}[1]{\colorbox{yellow}{#1}}

\newcommand{\bff}{\mathbf{f}}
\newcommand{\bfu}{\mathbf{u}}
\newcommand{\bfy}{\mathbf{y}}
\newcommand{\bfx}{\mathbf{x}}
\newcommand{\bft}{\mathbf{t}}
\newcommand{\bfk}{\mathbf{k}}
\newcommand{\bfmu}{\boldsymbol \mu}
\newcommand{\bfz}{\mathbf{0}}

\newcommand{\T}{{\top}}

\newcommand{\bfa}{\mathbf{a}}
\newcommand{\bb}{\beta^{-1}}
\newcommand{\la}{\left\langle}
\newcommand{\ra}{\right\rangle}
\newcommand{\vv}{\vartheta}

\newcommand{\intd}{\text{d}}

\date{}
\maketitle

\begin{abstract}
  High dimensional time series are endemic in applications of machine
  learning such as robotics (sensor data), computational biology (gene
  expression data), vision (video sequences) and graphics (motion
  capture data). Practical nonlinear probabilistic approaches to this
  data are required. In this paper we introduce the variational
  Gaussian process dynamical system. Our work builds on recent
  variational approximations for Gaussian process latent variable
  models to allow for nonlinear dimensionality reduction
  simultaneously with learning a dynamical prior in the latent
  space. The approach also allows for the appropriate dimensionality
  of the latent space to be automatically determined. We demonstrate
  the model on a human motion capture data set and a series of high
  resolution video sequences.
\end{abstract}

\section{Introduction}

Nonlinear probabilistic modeling of high dimensional time series data
is a key challenge for the machine learning community. A standard
approach is to simultaneously apply a nonlinear dimensionality
reduction to the data whilst governing the latent space with a
nonlinear temporal prior. The key difficulty for such approaches is
that analytic marginalization of the latent space is typically
intractable. Markov chain Monte Carlo approaches can also be
problematic as latent trajectories are strongly correlated making
efficient sampling a challenge. One promising approach to these time
series has been to extend the Gaussian process latent variable model
\cite{GPLVM,GPLVM2} with a dynamical prior for the latent space and
seek a maximum a posteriori (MAP) solution for the latent points
\cite{GPDM,Wang:gpdm08,hgplvm}. Ko and Fox \cite{GP-Based} further extend these
models for fully Bayesian filtering in a robotics setting. We refer to
this class of dynamical models based on the GP-LVM as Gaussian process
dynamical systems (GPDS). However, the use of a MAP approximation for
training these models presents key problems.  Firstly, since the
latent variables are not marginalised, the parameters of the dynamical
prior cannot be optimized without the risk of overfitting.
Further, the dimensionality of the latent space cannot be determined
by the model: adding further dimensions always increases the
likelihood of the data. In this paper we build on recent developments
in variational approximations for Gaussian processes
\cite{Titsias09,BayesianGPLVM} to introduce a variational Gaussian
process dynamical system (VGPDS) where latent variables are
approximately marginalized through optimization of a rigorous lower
bound on the marginal likelihood.  As well as providing a principled
approach to handling uncertainty in the latent space, this allows both
the parameters of the latent dynamical process and the dimensionality
of the latent space to be determined. The approximation enables the
application of our model to time series containing millions of
dimensions and thousands of time points. We illustrate this by
modeling human motion capture data and high dimensional video
sequences.

\section{The Model} 

Assume a multivariate times series dataset $\{\bfy_n,t_n\}_{n=1}^N$,
where $\bfy_n \in \mathbb{R}^D$ is a data vector observed at time $t_n
\in \mathbbm{R}_+$. We are especially interested in cases where each $\bfy_n$ is
a high dimensional vector and, therefore, we assume that there exists a low
dimensional manifold that governs the generation of the data.
Specifically, a {\em temporal} latent function $\bfx(t) \in
\mathbb{R}^Q$ (with $Q \ll D$), governs an intermediate {\em hidden}
layer when generating the data, and the $d$th feature from the
data vector $\bfy_n$ is then produced from $\bfx_n = \bfx(t_n)$
according to
\begin{equation}
\label{generative}
\mathit{y_{nd}} = f_d(\mathbf{x}_n) + \epsilon_{nd} \;, \;\;\; \epsilon_{nd} \sim \mathcal{N}(0, \beta^{-1}),
\end{equation}
where 
$f_d(\mathbf{x})$ is a latent mapping from the low dimensional space
to $d$th dimension of the observation space and $\beta$ is the inverse
variance of the white Gaussian noise.  We do not want to make strong
assumptions about the functional form of the latent functions $(\bfx,
\bff)$.\footnote{To simplify our notation, we often write $\bfx$
  instead of $\bfx(t)$ and $\bff$ instead of $\bff(\bfx)$. Later we
  also use a similar convention for the kernel functions by often
  writing them as $\mathit{k}_f$ and $\mathit{k}_x$.} Instead we
would like to infer them in a fully Bayesian non-parametric fashion
using Gaussian processes \cite{rasmussen-williams}.  Therefore, we
assume that $\bfx$ is a multivariate Gaussian process indexed by time
$t$ and $\bff$ is a different multivariate Gaussian process indexed by
$\bfx$, and we write
\begin{eqnarray}
x_q(t)  & \sim & \mathcal{GP}(0, k_x(t_i,t_j)), \ \ q=1,\ldots,Q, \\     
f_d(\bfx)  & \sim & \mathcal{GP}(0, k_f(\bfx_i,\bfx_j)), \ \ d=1,\ldots,D.
\label{eq:GPpriors}
\end{eqnarray}
Here, the individual components of the latent function $\bfx$ are
taken to be independent sample paths drawn from a Gaussian process
with covariance function $k_x(t_i,t_j)$. Similarly, the components of
$\bff$ are independent draws from a Gaussian process with covariance
function $k_f(\bfx_i,\bfx_j)$.  These covariance functions,
parametrized by parameters $\boldsymbol \theta_x$ and $\boldsymbol
\theta_f$ respectively, play very distinct roles in the model. More
precisely, $k_x$ determines the properties of each temporal latent
function $x_q(t)$. For instance, the use of an Ornstein-Uhlbeck
covariance function yields a Gauss-Markov process for $x_q(t)$, while
the squared-exponential kernel gives rise to very smooth and
non-Markovian processes. In our experiments, we will focus on the squared exponential
covariance function (RBF), the Matern $3/2$ which is only once
differentiable, and a periodic covariance function
\cite{rasmussen-williams, MacKay98} which can be used when data
exhibit strong periodicity. These kernel functions take the form:
\begin{eqnarray}
k_{x(\text{rbf})} \left( \mathit{t_i, t_j} \right) 
& = & \sigma_{\text{rbf}}^2 e^{- \frac{\left( t_i - t_j \right)^2}{\left(
      2l_t^2 \right)}}, 
\ \ k_{x(\text{mat})} \left( t_i, t_j \right) =  
\sigma_{\text{mat}}^2 \left( 1 + \frac{\sqrt{3} |t_i - t_j|}{l_t} \right)
		e^{\frac{ - \sqrt{3} |t_i - t_j|}{l_t} }, \nonumber \\
k_{x(\text{per})} \left( \mathit{t_i, t_j} \right) 
& = & 
	\sigma_{\text{per}}^2 e^{-\frac{1}{2} \frac{sin^2 \left( \frac{2
                \pi}{T} \left( t_i - t_j \right) \right) }{l_t} }. 
 \label{eq:temporalkernels}
\end{eqnarray}
The covariance function $k_f$ determines the properties of the latent
mapping $\bff$ that maps each low dimensional variable $\bfx_n$ to the
observed vector $\bfy_n$. We wish this mapping to be a non-linear but
smooth, and thus a suitable choice is the squared exponential
covariance function
\begin{align}
\mathit{k_f} \left( \mathbf{x}_i, \mathbf{x}_j \right) = {} &  
		\sigma_{ard}^2 e^{
			- \frac{1}{2} \sum_{q=1}^{Q}  w_q \left(
                          \mathit{x_{i,q} - x_{j,q}} \right) ^2 },
\label{rbfard}
\end{align}
which assumes a different scale $w_q$ for each latent dimension. This,
as in the variational Bayesian formulation of the GP-LVM
\cite{BayesianGPLVM}, enables an automatic relevance determination
procedure (ARD), i.e.\ it allows Bayesian training to ``switch off''
unnecessary dimensions by driving the values of the corresponding
scales to zero.

The matrix $\mathit{Y} \in \mathbb{R}^{N \times D}$ will collectively
denote all observed data so that its $n$th row corresponds to the data
point $\bfy_n$. Similarly, the matrix $F \in \mathbb{R}^{N \times D}$
will denote the mapping latent variables, i.e.\ $f_{nd} =f_d(\bfx_n)$,
associated with observations $Y$ from (\ref{generative}). Ana
usly,
$X \in \mathbb{R}^{N \times Q}$ will store all low dimensional latent
variables $x_{nq}=x_q(t_n)$. Further, we will refer to columns of
these matrices by the vectors $\bfy_d,\bff_d, \bfx_q \in
\mathbbm{R}^N$. Given the latent variables we assume independence over
the data features, and given time we assume independence over latent
dimensions to give
\begin{equation}
\label{joint}
p(Y,F, X | \bft) = p(Y|F) p(F|X) p(X |\bft)= 
 \prod_{d=1}^D  p(\mathbf{y}_d | \mathbf{f}_d) p(\mathbf{f}_d |
 \mathit{X}) \prod_{q=1}^Q p(\mathbf{x}_q | \mathbf{t}),
\end{equation}
where $\bft \in \mathbb{R}^N$ and $p(\mathbf{y}_d | \mathbf{f}_d)$ 
is a Gaussian likelihood function term defined from  (\ref{generative}). 
Further, $p(\mathbf{f}_d | \mathit{X})$ is a marginal GP prior 
such that 
\begin{equation}
\label{priorF}
p(\bff_d | \mathit{X}) = \mathcal{N}(\bff_d |\mathbf{0}, \mathit{K_{NN}}),
\end{equation}
where $\mathit{K}_{NN}= \mathit{k}_f(X,X)$ is the covariance matrix
defined by the kernel function $\mathit{k}_f$ and similarly 
$p(\bfx_q|\mathbf{t})$ is the marginal GP prior associated with 
the temporal function $x_q(t)$,  
\begin{equation}
p(\bfx_q|\mathbf{t}) = \mathcal{N} \left( \mathbf{x}_q | \mathbf{0},
  \mathit{K_t} \right),
\label{priorXgivenT}
\end{equation}
where $K_t = k_x(\bft,\bft)$ is the covariance matrix obtained by
evaluating the kernel function $\mathit{k}_x$ on the observed times
$\bft$. 
 
Bayesian inference using the above model poses a huge computational
challenge as, for instance, marginalization of the variables $X$, that
appear non-linearly inside the kernel matrix $K_{NN}$, is
troublesome. Practical approaches that have been considered until now
(e.g. \cite{hgplvm, GPDM}) marginalise out only $F$ and seek a MAP
solution for $X$.
%
 In the next section we describe how efficient variational 
approximations can be applied to marginalize $X$ by extending the 
framework of \cite{BayesianGPLVM}.

\subsection{Variational Bayesian training} 

The key difficulty with the Bayesian approach is propagating the prior
density $p(X|\bft)$ through the nonlinear mapping. This
mapping gives the expressive power to the model, but simultaneously
renders the associated marginal
likelihood,
\begin{equation}
\label{marginalLikelihood}
p(Y | \bft) =  \int p( Y | F) p(F | X ) p(X | \bft) \intd  X \intd F,
\end{equation}
intractable. We now invoke the variational Bayesian methodology to
approximate the integral. Following a standard procedure
\cite{bishop}, we introduce a variational distribution $q(\Theta)$ and
compute the Jensen's lower bound $\mathcal{F}_v$ on the logarithm of
\eqref{marginalLikelihood},
\begin{align}
\mathcal{F}_v(q, \boldsymbol \theta) = {}& \int q(\mathit{\Theta}) \log 
		\frac{ p(Y | F) p(F | X ) p(\mathit{X} | \mathbf{t})}
			 {q(\mathit{\Theta})}  \intd  X \intd F,
		 \label{jensens1}
\end{align}
where $\boldsymbol \theta$ denotes the model's parameters.  However,
the above form of the lower bound is problematic becauce $X$ (in the
GP term $p(F|X)$) appears non-linearly inside the kernel matrix
$K_{NN}$ making the integration over $X$ difficult. As shown in
\cite{BayesianGPLVM}, this intractability is removed by applying the
``data augmentation'' principle.  More precisely, we augment the joint
probability model in (\ref{joint}) by including $M$ extra samples of
the GP latent mapping $\bff$, known as inducing points, so that
$\bfu_m \in \mathbb{R}^D$ is such a sample. The inducing points are
evaluated at a set of pseudo-inputs $\tilde{X} \in \mathbb{R}^{M
  \times Q}$. 
The augmented joint probability density takes the form
\begin{equation}
 \label{augmentedJoint}
p(Y,F, U,X,\tilde{X} | \bft) = \prod_{d=1}^D p(\mathbf{y}_d | \mathbf{f}_d) p(\mathbf{f}_d | \mathbf{u}_d, \mathit{X})
p(\bfu_d | \tilde{X})  p(X | \mathbf{t}),
\end{equation}
where $p(\bfu_d | \tilde{X})$ is a zero-mean Gaussian with a
covariance matrix $K_{MM}$ constructed using the same function as for
the GP prior \eqref{priorF}. By dropping $\tilde{X}$ from our
expressions, we write the augmented GP prior analytically (see
\cite{rasmussen-williams}) as
\begin{equation}
 \label{priorF2}
p(\bff_d | \bfu_d, X) =  \mathcal{N}  \left( \bff_d | K_{NM} K_{MM}^{-1} \bfu_d , K_{NN} - K_{NM} K_{MM}^{-1} K_{MN} \right).
\end{equation}
A key result in \cite{BayesianGPLVM}
 is that a tractable lower bound 
(computed analogously to (\ref{jensens1})) can be obtained through the variational density
\begin{equation}
\label{varDistr}
q(\mathit{\Theta}) = q(F, U,X) = q(F | U, X) q(U) q(X) = \prod_{d=1}^D p(\bff_d | \bfu_d, X )q(\bfu_d) q(X),
\end{equation}
where $q(X) = \prod_{q=1}^Q \mathcal{N} \left( \bfx_q | \bfmu_q, S_q
\right)$ and $q(\bfu_d)$ is an arbitrary variational distribution. 
Titsias and Lawrence \cite{BayesianGPLVM} assume full independence for
$q(X)$ and the variational covariances are diagonal matrices.
Here, in contrast, 
the posterior over the latent variables will have strong correlations, 
so $S_q$ is taken to be a $N \times N$ full covariance
matrix. Optimization of the variational lower bound provides 
an approximation to the true posterior $p(X|Y)$ by $q(X)$.
In the augmented probability model, the ``difficult'' term $p(F | X)$
appearing in \eqref{jensens1} is now replaced with \eqref{priorF2}
and, eventually, it cancels out with the first factor of the
variational distribution \eqref{varDistr} so that $F$ can be
marginalised out analytically.
%
%
Given the above and after breaking the logarithm in \eqref{jensens1},
we obtain the final form of the lower bound (see supplementary
material for more details)
\begin{equation}
\label{jensens}
\mathcal{F}_v(q, \boldsymbol \theta) = 
\hat{\mathcal{F}}_v - \text{KL}(q(X) \parallel p(X|\bft)),
\end{equation}
with $\hat{\mathcal{F}}_v =\int q(X) \log p( Y | F ) p( F | X) \,
\mathrm{d} X \intd F$. Both terms in \eqref{jensens} are now
tractable. Note that the first of the above terms involves the data while
the second one only involves the prior. All the information regarding
data point correlations is captured in the $\text{KL}$ term and the
connection with the observations comes through the variational
distribution. Therefore, the first term in \eqref{jensens} has the
same analytical solution as the one derived in \cite{BayesianGPLVM}.
%
\eqref{jensens} can be maximized by using gradient-based
methods\footnote{See supplementary material for more detailed
  derivation of \eqref{jensens} and for the equations for the
  gradients.}. However, when not factorizing $q(X)$ across data points
yields $O(N^2)$ variational parameters to optimize. 
This issue is addressed in the next section.

\subsection{Reparametrization and Optimization \label{optimisation}} 

The optimization involves the model parameters $\boldsymbol \theta =
(\beta, \boldsymbol \theta_f, \boldsymbol \theta_t)$, the variational
parameters $\{ \bfmu_q, S_q \}_{q=1}^Q$ from $q(X)$ and the inducing
points\footnote{We will use the term ``variational parameters'' to
  refer only to the parameters of $q(X)$ although the inducing points
  are also variational parameters.} $\tilde{X}$.

Optimization of the variational parameters appears challenging, due to
their large number and the correlations between them. However, by
reparametrizing our $O \left( N^2 \right)$ variational parameters
according to the framework described in
\cite{OpperFixedPointCovariance} we can obtain a set of $O(N)$ less
correlated variational parameters. Specifically, we first take the
derivatives of the variational bound \eqref{jensens} w.r.t. $S_q$ and
$\bfmu_q$ and set them to zero, to find the stationary
points,
\begin{equation}
S_q = \left( \mathit{K}_t^{-1} + \Lambda_q \right)^{-1} \;\;\; \text{and}  \;\;\;  \boldsymbol \mu_q = K_t \bar{\boldsymbol \mu}_q, \label{SFixedPointQ}
\end{equation}
where $\Lambda_q = - 2\frac{\vartheta \mathit{\hat{F}_v(q, \boldsymbol
    \theta)}}{\vartheta \mathit{S_q}}$ is a $N \times N$ diagonal,
positive matrix and $\bar{\boldsymbol \mu}_q = \frac{\vartheta
  \hat{F}_v}{\vartheta \boldsymbol \mu_q}$ is a $N-$dimensional
vector.
The above stationary conditions tell us that, since $S_q$ depends on a
diagonal matrix $\Lambda_q$, we can reparametrize it using only the
$N-$dimensional diagonal of that matrix, denoted by $\boldsymbol
\lambda_q$.  Then, we can optimise the $2 (Q \times N)$ parameters $(
\boldsymbol \lambda_q$, $\bar{\bfmu}_q)$ and obtain the original
parameters using \eqref{SFixedPointQ}.

\subsection{Learning from Multiple Sequences \label{sequences}}

Our objective is to model multivariate time series. A given data set
may consist of a group of independent observed sequences, each with
a different length (e.g.\ in human motion capture data several walks
from a subject). Let, for example, the dataset be a group of
$S$ independent sequences  $\left( Y^{(1)}, ..., Y^{(S)} \right)$. We would like our model to capture the underlying
commonality of these data. We handle this by allowing a different temporal latent function for each of the independent
sequences, so that $X^{(s)}$ is the set of latent variables corresponding to the sequence $s$.
%
These sets are a priori assumed to be independent since they correspond to separate sequences, i.e.\ $p\left( X^{(1)}, X^{(2)}, ..., X^{(S)} \right) = \prod_{s=1}^S p(X^{(s)})$, where we dropped the
conditioning on time for simplicity.
%
This factorisation leads to a block-diagonal structure for the time covariance matrix $K_t$, where each block corresponds to one sequenece.
 In this setting, each block of observations $Y^{(s)}$ is generated from its corresponding $X^{(s)}$
according to $Y^{(s)} = F^{(s)} + \boldsymbol \epsilon$, where the latent function which governs this mapping is shared across all sequences and 
$\boldsymbol \epsilon$ is Gaussian noise.

\section{Predictions} 

Our algorithm models the temporal evolution of a dynamical system. It
should be capable of generating completely new sequences or
reconstructing missing observations from partially observed data. For
generating novel sequence given training data the model requires a
time vector $\bft_*$ as input and computes a density $p(Y_* | Y, \bft,
\bft_*)$. For reconstruction of partially observed data the time stamp
information is additionally accompanied by a partially observed
sequence $Y_*^{p} \in \mathbb{R}^{N_* \times D_p}$ from the whole $Y_*
= (Y_*^{p}, Y_*^{m})$, where $p$ and $m$ are set of indices indicating
the present (i.e. observed) and missing dimensions of $Y_*$
respectively, so that $p \cup m= \{1,\ldots,D\}$.  We reconstruct the
missing dimensions by computing the Bayesian predictive distribution
$p(Y_*^{m}| Y_*^{p}, Y, \bft_*, \bft)$. The predictive densities can
also be used as estimators for tasks like generative Bayesian
classification.
 Whilst time stamp information is always provided, in the
next section we drop its dependence to avoid notational clutter.

\subsection{Predictions Given Only the Test Time Points \label{unobservedData}}
To approximate the predictive density, we will need to introduce the underlying latent function values $F_* \in \mathbb{R}^{N_* \times D}$ (the noisy-free version of $Y_*$) and the latent variables $X_* \in \mathbb{R}^{N_* \times Q}$. We  write the predictive density as
\begin{eqnarray}
p(Y_* | Y) & = & \int p(Y_*, F_*, X_*| Y_*, Y) \intd  F_* \intd  X_* =  \int p(Y_* | F_*)  p(F_*|X_*, Y) p(X_*|  Y) \intd  F_* \intd  X_* .
\label{eq:predictive1}
\end{eqnarray}
The term $p(F_* |X_*, Y)$ is approximated by the variational distribution
\begin{eqnarray}
q(F_*|X_*) & = & \int \prod_{d \in D} p(\bff_{*,d} | \bfu_d, X_*)  q(\bfu_d) d \bfu_d 
	    = \prod_{d \in D} q(\bff_{*,d} | X_*)  ,
\end{eqnarray}
where $q(\bff_{*,d} | X_*)$ is a Gaussian that can be computed analytically,
since in our variational framework the optimal setting for $q(\bfu_d)$ is also found to be a Gaussian (see suppl. material for complete forms).
As for the term $p(X_*| Y)$ in eq. (\ref{eq:predictive1}), it is approximated by
a Gaussian variational distribution $q(X_*)$,
\begin{align}
 q(X_*) = \prod_{q=1}^Q q(\bfx_{*,q}) = 
\prod_{q=1}^Q   \int  p(\bfx_{*,q} | \bfx_q) q(\bfx_q) \intd \bfx_q = \prod_{q=1}^Q \la  p(\bfx_{*,q} | \bfx_q) \ra_{q(\bfx_q)} ,\label{qxstar}
\end{align}
where $p(\bfx_{*,q} | \bfx_q)$ is a Gaussian found from the conditional GP prior
(see \cite{rasmussen-williams}) and $q(X)$ is also Gaussian. We can, thus, work out analytically the mean and variance 
for \eqref{qxstar}, which turn out to be:
\begin{align}
 \mu_{x_{*,q}} = {}& K_{*N} \bar{\mu}_q \\
  \text{var}(x_{*,q}) = {}& K_{**} - K_{*N} (K_t + \Lambda_q^{-1})^{-1} K_{N*}.
\end{align}
where $K_{*N} = k_x(\bft_*, \bft)$, $K_{*N} = K_{*N}^\T$ and $K_{**} = k_x(\bft_*, \bft_*)$. Notice that these equations have
exactly the same form as found in standard GP regression problems.
Once we have analytic forms for the posteriors in \eqref{eq:predictive1}, the predictive density is approximated as
\begin{align} 
p(Y_*| Y) {}& =  \int p(Y_*| F_*)  q(F_*|X_*) q(X_*) \intd F_* \intd X_* = \int p(Y_* | F_*) \la q(F_* | X_*) \ra_{q(X_*)} \intd F_* , \label{eq:predictive2}
\end{align}
which is a non-Gaussian integral that cannot be computed analytically. However, following the same argument as in \cite{rasmussen-williams, Girard03gaussianprocess}, we can
calculate analytically its mean and covariance:
%
\begin{align}
 \mathbb{E}(F_*) ={}&  B^\T \Psi_1^* \label{meanFstar} \\
 \text{Cov}(F_*) ={}& B^\T \left( \Psi_2^* - \Psi_1^* (\Psi_1^*)^\T \right) B + \Psi_0^* I - \text{Tr} \left[ \left( K_{MM}^{-1} - \left( K_{MM} + \beta \Psi_2 \right)^{-1} \right) \Psi_2^* \right] I,
\end{align}
where $B = \beta \left( K_{MM} + \beta \Psi_2 \right)^{-1} \Psi_1^\T
Y$, $\Psi_0^* = \la k_f(X_*, X_*) \ra$, $\Psi_1^* = \la K_{M*} \ra$
and $\Psi_2^* = \la K_{M*} K_{*M} \ra$. All expectations are taken
w.r.t. $q(X_*)$ and can be calculated analytically, while $K_{M*}$
denotes the cross-covariance matrix between the training inducing
inputs $\tilde{X}$ and $X_*$. The $\Psi$ quantities are calculated analytically (see suppl. material). Finally, since $Y_*$ is just a noisy version of
$F_*$, the mean and covariance of \eqref{eq:predictive2} is just
computed as: $\mathbb{E}(Y_*) = \mathbb{E}(F_*)$ and $\text{Cov}(Y_*)
= \text{Cov}(F_*) + \beta^{-1} I_{N_*}$.

\subsection{Predictions Given the Test Time Points and Partially Observed Outputs}

The expression for the predictive density $p(Y_*^m | Y_*^p, Y)$ is similar to  \eqref{eq:predictive1},
\begin{eqnarray}
p(Y_*^m | Y_*^p, Y) & = & \int p(Y_*^m | F_*^m)  p(F_*^m|X_*, Y_*^p, Y) p(X_*|  Y_*^p, Y) \intd  F_*^m \intd  X_* ,
\label{eq:predictive3}
\end{eqnarray}
and is analytically intractable. 
To obtain an approximation, we firstly need to apply variational inference and approximate $p(X_* | Y_*^p, Y)$ with a Gaussian distribution.
This requires the optimisation of a new variational lower bound that accounts for the contribution of the partially observed data $Y_*^p$. 
This lower bound approximates the true marginal likelihood $p(Y_*^p, Y)$ and has exactly analogous form with the lower bound computed only on the training data $Y$. 
Moreover, the variational optimisation requires the definition of the variational distribution $q(X_*,X)$ which needs to be optimised and is fully correlated across $X$ and $X_*$. 
After the optimisation, the approximation to the true posterior  $p(X_* | Y_*^p, Y)$ is given from the marginal $q(X_*)$. 
 A much faster but less
accurate method would be to decouple the test from the training latent
variables by imposing the factorisation $q(X_*, X) = q(X)
q(X_*)$. This is not used, however, in our current implementation.

\section{Handling Very High Dimensional Datasets}

Our variational framework avoids the typical cubic complexity of
Gaussian processes allowing relatively large training sets (thousands
of time points, $N$). Further, the model scales only linearly with the
number of dimensions $D$. Specifically, the number of dimensions only
matters when performing calculations involving the data matrix $Y$. In
the final form of the lower bound (and consequently in all of the
derived quantities, such as gradients) this matrix only appears in the
form $Y Y^\T$ which can be precomputed. This means that, when $N \ll
D$, we can calculate $Y Y^\T$ only once and then substitute $Y$ with
the SVD (or Cholesky decomposition) of $Y Y^\T$. In this way, we can
work with an $N \times N$ instead of an $N \times D$
matrix. Practically speaking, this allows us to work with data sets
involving millions of features. In our experiments we model directly
the pixels of HD quality video, exploiting this trick.

\begin{matlab}


clear all
close all

retrainModels = 0;

rePredict = 0;


save 'opts.mat' 'retrainModels' 'rePredict'
system('mkdir ../diagrams');

clear; load opts
fprintf(1,'\n\n#-----  OCEAN DEMO ----#\n');
dataSetName = 'ocean';
experimentNo=60;
indPoints=-1; latentDim=20;
fixedBetaIters=200; reconstrIters = 2000;
itNo=[1000 2000 5000 8000 4000];
dynamicKern={'rbf','white','bias'};
whiteVar = 0.1;  vardistCovarsMult=1.7;
dataSetSplit = 'randomBlocks';
if retrainModels
    demHighDimVargplvm3
elseif rePredict
    demHighDimVargplvmTrained
else
    load demOceanVargplvm60Pred
    demHighDimVargplvmLoadPred
end
fr=reshape(Varmu(27,:),height,width); imagesc(fr); colormap('gray'); 
print -depsc oceanGpdsframe27.eps; system('epstopdf oceanGpdsframe27.eps');
fr=reshape(NNmuPartBest(27,:),height,width); imagesc(fr); colormap('gray'); 
print -depsc oceanNNframe27.eps; system('epstopdf oceanNNframe27.eps');

clear; load opts
fprintf(1,'\n\n#-----  MISSA DEMO ----#\n');
experimentNo = 59;
dataSetName = 'missa';
indPoints = -1; latentDim=25;
fixedBetaIters=50; reconstrIters = 4000;
itNo=[1000 2000 5000 8000 2000]; 
dynamicKern={'matern32','white','bias'};
vardistCovarsMult=1.6;
dataSetSplit = 'blocks';
blockSize = 4; whiteVar = 0.1;
msk = [48 63 78 86 96 111 118];
if retrainModels
    demHighDimVargplvm3
elseif rePredict
    demHighDimVargplvmTrained
else
    load demMissaVargplvm59Pred
    demHighDimVargplvmLoadPred
end

fr=reshape(Varmu(46,:),height,width); imagesc(fr); colormap('gray'); 
print -depsc missaGpdsframe46.eps; system('epstopdf missaGpdsframe46.eps');
fr=reshape(YtsOriginal(46,:),height,width); imagesc(fr); colormap('gray'); 
print -depsc missaYtsOrigframe46.eps; system('epstopdf missaYtsOrigframe46.eps');
fr=reshape(NNmuPartBest(46,:),height,width); imagesc(fr); colormap('gray'); 
print -depsc missaNNframe46.eps; system('epstopdf missaNNframe46.eps');
fr=reshape(Yts(17,:),height,width); imagesc(fr); colormap('gray'); 
print -depsc missaGpdsPredFrame17_part1.eps; system('epstopdf missaGpdsPredFrame17_part1.eps');
fr=reshape(Varmu(17,:),height,width); imagesc(fr); colormap('gray');
print -depsc missaGpdsPredFrame17_part2.eps; system('epstopdf missaGpdsPredFrame17_part2.eps');


clear; load opts
fprintf(1,'\n\n#-----  DOG DEMO: Generation ----#\n');
dataSetName = 'dog';
experimentNo=61;
indPoints=-1; latentDim=35;
fixedBetaIters=400;
reconstrIters = 1; 
itNo=[1000 1000 1000 1000 1000 1000 500 500 500 500 1000 1000 1000 1000 1000 1000 1000 500 500]; 
periodicPeriod = 4.3983; 
dynamicKern={'rbfperiodic','whitefixed','bias','rbf'};
vardistCovarsMult=0.8;
whiteVar = 1e-6;
dataToKeep = 60; dataSetSplit = 'custom';
indTr = [1:60];
indTs = 60; 
learnSecondVariance = 0;
if retrainModels
    demHighDimVargplvm3
end

clear; load opts; close all 
dataSetName = 'dog'; 
experimentNo=61; dataToKeep = 60; dataSetSplit = 'custom';
indTr = [1:60]; indTs = 60;
futurePred = 40; doSampling = 0; demHighDimVargplvmTrained

bar(prunedModelInit.kern.comp{1}.inputScales)
print -depsc dog_scalesInit.eps; system('epstopdf dog_scalesInit.eps');
bar(model.kern.comp{1}.inputScales)
print -depsc dog_scalesOpt.eps; system('epstopdf dog_scalesOpt.eps');

fr=reshape(Ytr(end,:),height,width); imagesc(fr); colormap('gray'); 
print -depsc dogGeneration_lastOfTraining.eps; system('epstopdf dogGeneration_lastOfTraining.eps');
fr=reshape(Varmu2(1,:),height,width); imagesc(fr); colormap('gray');  
print -depsc dogGeneration_firstOfTest.eps; system('epstopdf dogGeneration_firstOfTest.eps');
fr=reshape(Varmu2(13,:),height,width); imagesc(fr); colormap('gray'); 
print -depsc dogGeneration_frame14.eps; system('epstopdf dogGeneration_frame14.eps');



clear; load opts
fprintf(1,'\n\n#-----  DOG DEMO: Reconstruction ----#\n');
dataSetName = 'dog';
experimentNo=65;
indPoints=-1; latentDim=35;
fixedBetaIters=400;
reconstrIters = 2;
itNo=[1000 1000 1000 1000 1000 1000 500 500 500 500 1000 1000 1000 1000 1000 1000 1000 500 500]; 
periodicPeriod = 2.8840;
dynamicKern={'rbfperiodic','whitefixed','bias','rbf'};
vardistCovarsMult=0.8;
whiteVar = 1e-6;
dataSetSplit = 'custom';
indTr = 1:54;
indTs = 55:61;
learnSecondVariance = 0;
if retrainModels
    demHighDimVargplvm3
end


clear; load opts
dataSetName = 'dog';
experimentNo=65;
dataSetSplit = 'custom';
indTr = 1:54; indTs = 55:61;
predWithMs = 1; 
reconstrIters = 18000; 
doSampling = 0;
if rePredict
    demHighDimVargplvmTrained
else
    load demDogVargplvm65Pred
    demHighDimVargplvmLoadPred
end
fr=reshape(Varmu(5,:),height,width); imagesc(fr); colormap('gray'); 
print -depsc supplDogPredGpds5.eps; system('epstopdf supplDogPredGpds5.eps');
fr=reshape(Yts(5,:),height,width); imagesc(fr); colormap('gray'); 
print -depsc supplDogPredYts5.eps; system('epstopdf supplDogPredYts5.eps');
fr=reshape(Varmu(6,:),height,width); imagesc(fr); colormap('gray'); 
print -depsc supplDogPredGpds6.eps; system('epstopdf supplDogPredGpds6.eps');
fr=reshape(Yts(6,:),height,width); imagesc(fr); colormap('gray'); 
print -depsc supplDogPredYts6.eps; system('epstopdf supplDogPredYts6.eps');


clear; load opts
fprintf(1,'\n\n#-----  CMU DEMO: Rbf ----#\n');
experimentNo=34; 
itNo = [300 300 400 200 200 300 400 400];
dynamicKern = {'rbf', 'white', 'bias'};
vardistCovarsMult = 0.152;
if retrainModels 
    if rePredict
        doReconstr = 1;
    else
        doReconstr=0;
    end
    demCmu35gplvmVargplvm3;
elseif rePredict
    demCmu35vargplvmReconstructTaylor
end
predictPart = 'Legs';  plotRange = [];
demCmu35VargplvmPlotsScaled
fprintf(1,'# VGPDS RBF error on Legs reconstr:');
errStruct

predictPart = 'Body';  plotRange = [];
demCmu35VargplvmPlotsScaled
fprintf(1,'# VGPDS RBF error on Body reconstr:');
errStruct

bar(model.kern.comp{1}.inputScales);
print -depsc supplMocapScalesRbf.eps; system('epstopdf supplMocapScalesRbf.eps');

clear; load opts
fprintf(1,'\n\n#-----  CMU DEMO: Matern32 ----#\n');
experimentNo=33; 
itNo = [300 300 400 200 200 300 400 400];
dynamicKern = {'matern32', 'white', 'bias'};
vardistCovarsMult = 0.24;
if retrainModels 
    if rePredict
        doReconstr = 1;
    else
        doReconstr=0;
    end
    demCmu35gplvmVargplvm3;
elseif rePredict
    demCmu35vargplvmReconstructTaylor
end
predictPart = 'Legs'; plotRange = 10;
demCmu35VargplvmPlotsScaled
print -depsc supplMocapLeg5GpdsMatern.eps; system('epstopdf supplMocapLeg5GpdsMatern.eps');
fprintf(1,'# VGPDS Matern error on Legs reconstr:');
errStruct

predictPart = 'Body'; plotRange = 28;
demCmu35VargplvmPlotsScaled
print -depsc supplMocapBody28GpdsMatern.eps; system('epstopdf supplMocapBody28GpdsMatern.eps');
fprintf(1,'# VGPDS Matern error on Body reconstr:');
errStruct
close all
bar(model.kern.comp{1}.inputScales);
print -depsc supplMocapScalesMatern.eps; system('epstopdf supplMocapScalesMatern.eps');

fprintf(1,'\n\n#---- FINISHED reproducing plots and results!! \n');
delete opts.mat

a = ver('octave');
if length(a) == 0
  a = ver('matlab');
end
fid = fopen('vers.tex', 'w');
fprintf(fid, [a.Name ' version ' a.Version]);
fclose(fid);

fid = fopen('computer.tex', 'w');
fprintf(fid, ['\\verb+' computer '+']);
fclose(fid);

fid = fopen('date.tex', 'w');
fprintf(fid, datestr(now, 'dd/mm/yyyy'));
fclose(fid);

\end{matlab}

\section{Experiments}

We consider two different types of high dimensional time series, a
human motion capture data set consisting of different walks and high
resolution video sequences. The experiments are intended to explore
the various properties of the model and to evaluate its performance in
different tasks (prediction, reconstruction, generation of data). 
Matlab source code for repeating the following experiments is available
on-line from 
\verb|http://staffwww.dcs.shef.ac.uk/people/N.Lawrence/vargplvm/|.

\subsection{Human Motion Capture Data}

We followed \cite{Taylor,gplvmLarger} in considering motion capture
data of walks and runs taken from subject 35 in the CMU motion capture
database. We treated each motion as an independent sequence.  The data
set was constructed and preprocessed as described in
\cite{gplvmLarger}. This results in 2,613 separate 59-dimensional
frames split into 31 training sequences with an average length of 84
frames each.

The model is jointly trained, as explained in section \ref{sequences},
on both walks and runs, i.e. the algorithm learns a common latent
space for these motions. At test time we investigate the ability of
the model to reconstruct test data from a previously unseen sequence
given partial information for the test targets. This is tested once by
providing only the dimensions which correspond to the body of the
subject and once by providing those that correspond to the legs.
We compare with results in \cite{gplvmLarger}, which used MAP
approximations for the dynamical models, and against nearest
neighbour. We can also indirectly compare with the binary latent
variable model (BLV) of \cite{Taylor} which used a slightly different
data preprocessing. We assess the performance using the cumulative
error per joint in the scaled space defined in \cite{Taylor} and by
the root mean square error in the angle space suggested by
\cite{gplvmLarger}. Our model was initialized with nine latent
dimensions. We performed two runs, once using the Matern covariance
function for the dynamical prior and once using the RBF. From table
\ref{motionCaptureTable} we see that the variational Gaussian process
dynamical system considerably outperforms the other approaches.
The appropriate latent space dimensionality for the data was
automatically inferred by our models. 
The model which employed an RBF covariance to govern the dynamics retained four dimensions,
whereas the model that used the Matern kept only three.
%
%
The other latent dimensions were completely switched off by the ARD
parameters.  The best performance for the legs and the body
reconstruction was achieved by the VGPDS model that used the Matern
and the RBF covariance function respectively.

\begin{table}[h]
\caption{
\small{
Errors obtained for the motion capture dataset considering nearest neighbour in the angle space (NN) and in the scaled space(NN sc.), GPLVM, BLV and VGPDS. CL / CB are the leg and body datasets as preprocessed in \cite{Taylor}, L and B the corresponding datasets from \cite{gplvmLarger}. SC corresponds to the error in the scaled space, as in Taylor et al. while RA is the error in the angle space. The best error per column is in bold. }}
\label{motionCaptureTable}
\begin{center}
\begin{tabular}{c||c|c|c|c|c|c}
Data & CL & CB & L & L & B & B \\  \hline
Error Type & SC & SC & SC & RA & SC & RA \\
\hline \hline
BLV 			       & 11.7 & \textbf{8.8} & - & - & - & - \\  \hline
NN sc.   		       & 22.2 & \textbf{20.5} & - & - & - & - \\ \hline
GPLVM (Q = 3)	       & - & - & 11.4 & 3.40 & 16.9 & 2.49 \\ \hline
GPLVM (Q = 4)	       & - & - & 9.7  & 3.38 & 20.7 & 2.72 \\ \hline
GPLVM (Q = 5)	       & - & - & 13.4 & 4.25 & 23.4 & 2.78 \\ \hline
NN sc.  		       & - & - & 13.5 & 4.44 & 20.8 & 2.62 \\ \hline
NN 		 		       & - & - & 14.0 & 4.11 & 30.9 & 3.20 \\ \hline
VGPDS (RBF)        & - & - & 8.19 & 3.57 & \textbf{10.73} & \textbf{1.90} \\ \hline
VGPDS (Matern 3/2) & - & - & \textbf{6.99} & \textbf{2.88} & 14.22 & 2.23 \\
\end{tabular}
\end{center}
\end{table}

\subsection{Modeling Raw High Dimensional Video Sequences}

For our second set of experiments we considered video sequences. Such
sequences are typically preprocessed before modeling to extract
informative features and reduce the dimensionality of the
problem. Here we work directly with the raw pixel values to
demonstrate the ability of the VGPDS to model data with a vast number
of features. This also allows us to directly sample video from the
learned model.
\par Firstly, we used the
model to reconstruct partially observed frames from test video
sequences\footnote{`Missa' dataset: cipr.rpi.edu. `Ocean': cogfilms.com. `Dog': fitfurlife.com. See details in supplementary. The logo appearing in the `dog' images in the experiments that follow, has been added with post-processing.}. For the first video discussed here we gave as partial information approximately 
50\% of the pixels while for the other two we gave approximately 40\% of the pixels on each frame.
The mean squared error per pixel was measured to compare
with the $k-$nearest neighbour (NN) method, for $k \in (1,..,5)$ (we
only present the error achieved for the best choice of $k$ in each
case). The datasets considered are the following: firstly, the `Missa'
dataset, a standard benchmark used in image
processing. This is 103,680-dimensional video, showing a woman talking
for 150 frames. The data is challenging as there are translations in
the pixel space. We also considered an HD video of dimensionality $9
\times 10^5$ that shows an artificially created scene of ocean waves
as well as a $230,400-$dimensional video showing
a dog running for $60$ frames. The later is approximately periodic in
nature, containing several paces from the dog. For the first two
videos we used the Matern and RBF kernel respectively to model the
dynamics and interpolated to reconstruct blocks of frames chosen from
the whole sequence. For the `dog' dataset we constructed a compound
kernel $k_x = k_{x(\text{rbf})} + k_{x(\text{periodic})}$, where the
RBF term is employed to capture any divergence from the approximately
periodic pattern. We then used our model to reconstruct the last 7
frames extrapolating beyond the original video. As can be seen in
table \ref{videoResultsTable}, our method outperformed NN in all
cases. The results are also demonstrated visually in figure
\ref{fig:video1} and the reconstructed videos are available in the supplementary material.

\begin{table}[h]
\caption{
\small{
The mean squared error per pixel for VGPDS and NN for the three datasets (measured only in the missing inputs). The number of latent dimensions selected by our model is in parenthesis. 
} }
\label{videoResultsTable}
\begin{center}
\begin{tabular}{c||l|l|l}
  & Missa & Ocean & Dog \\
\hline \hline
VGPDS  & 2.52 ($Q = 12$) & 9.36 ($Q = 9$)  & 4.01 ($Q = 6$) \\  \hline
NN  & 2.63 & 9.53 & 4.15 \\
\end{tabular}
\end{center}
\end{table}

\begin{figure}[ht]
\begin{center}
\subfigure[]{
\includegraphics[width=0.24\textwidth]{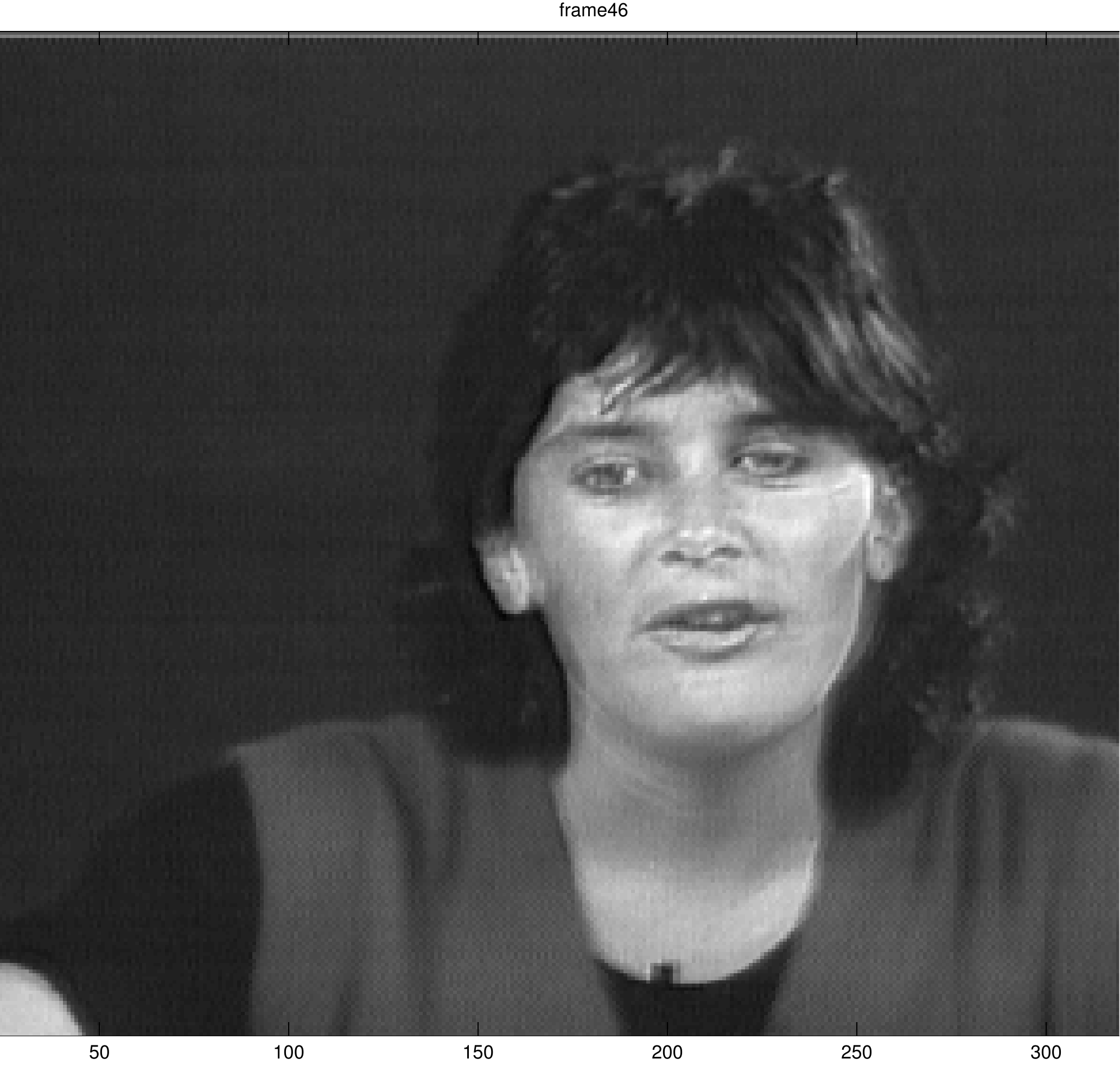}
	\label{fig:missa1}
}
\subfigure[]{
	\includegraphics[width=0.24\textwidth]{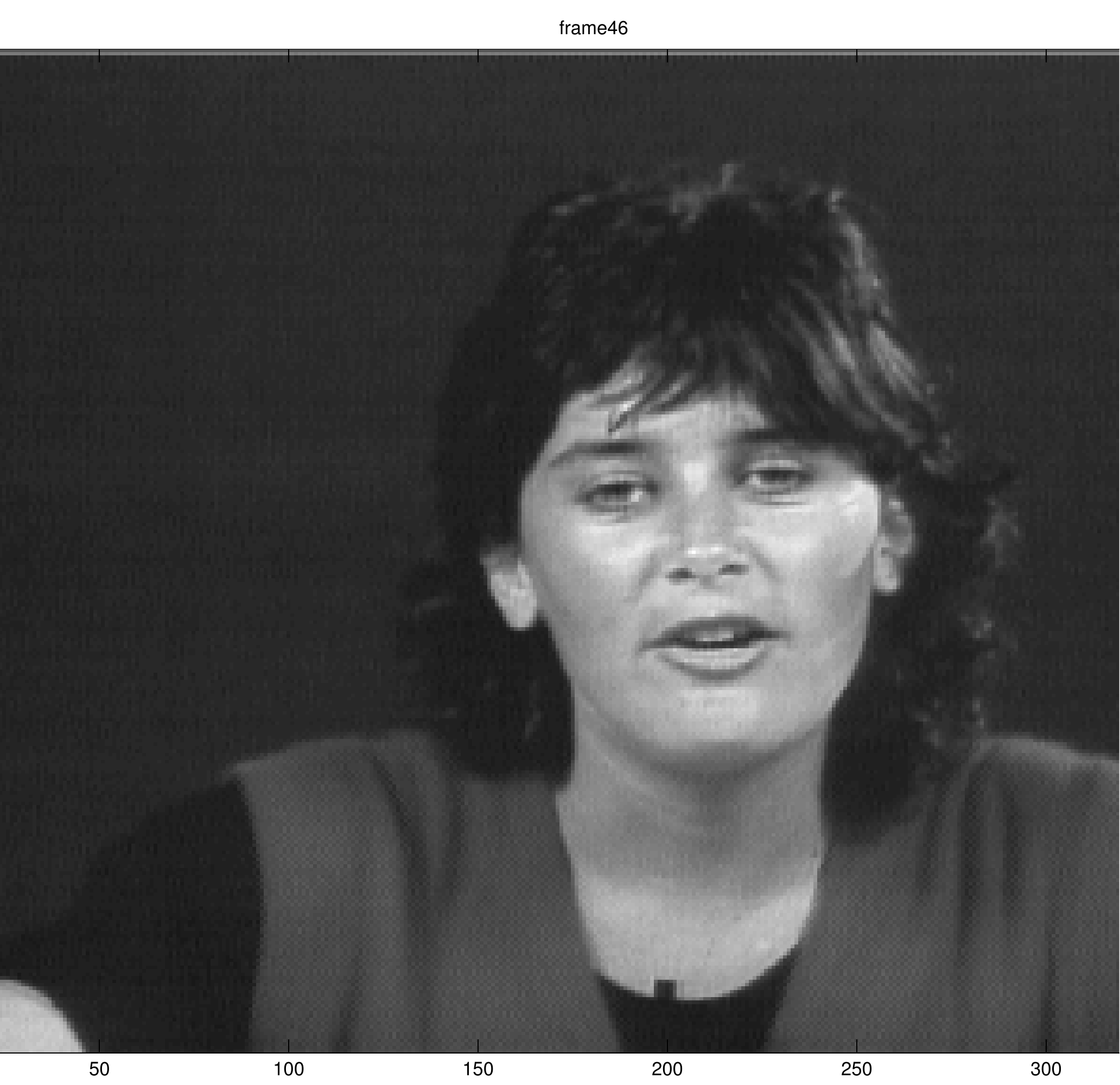}
	\label{fig:missa2}
}
\subfigure[]{
	\includegraphics[width=0.24\textwidth]{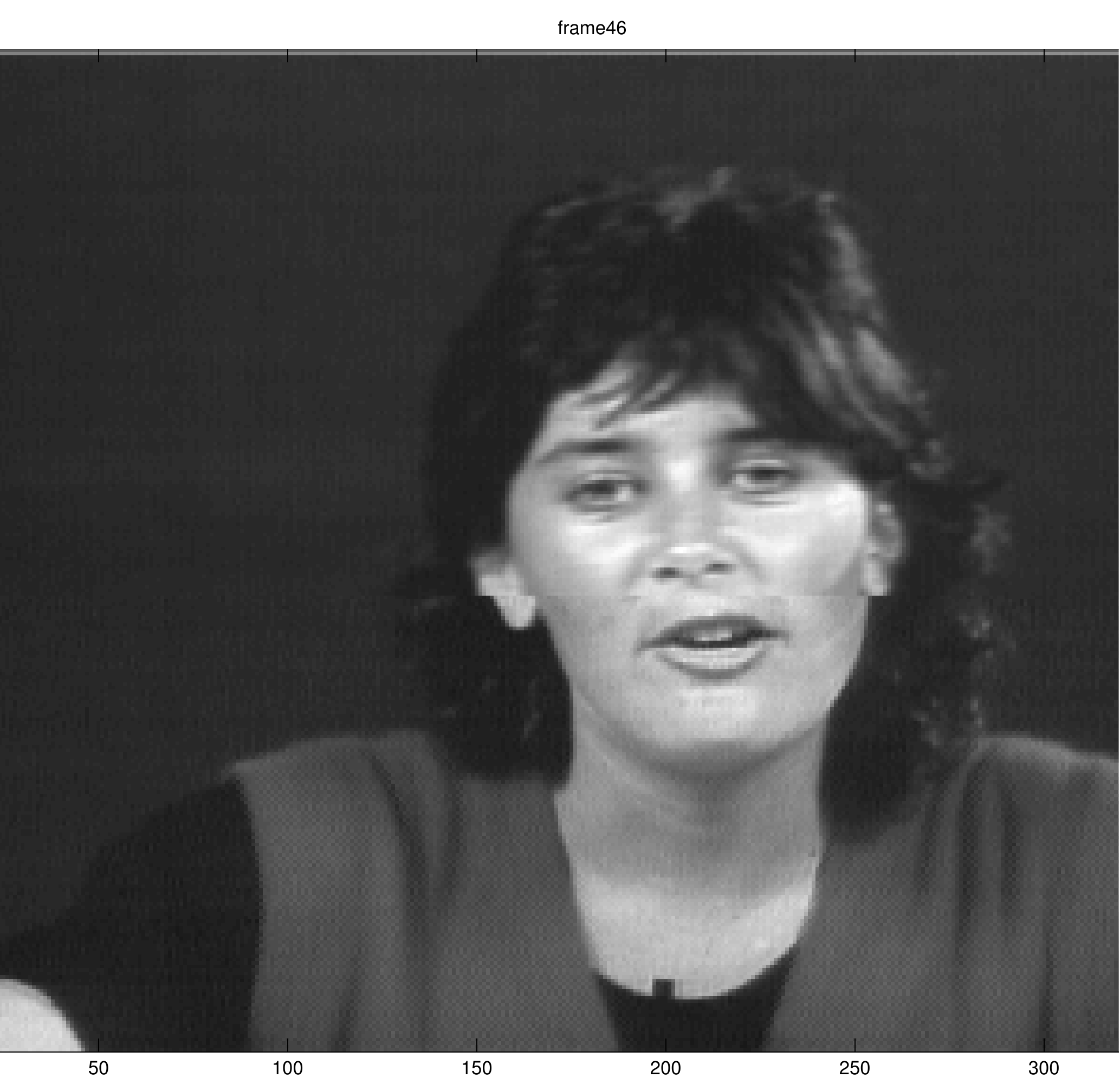}
	\label{fig:missa3}
}
\subfigure[]{
	\includegraphics[width=0.12\textwidth]{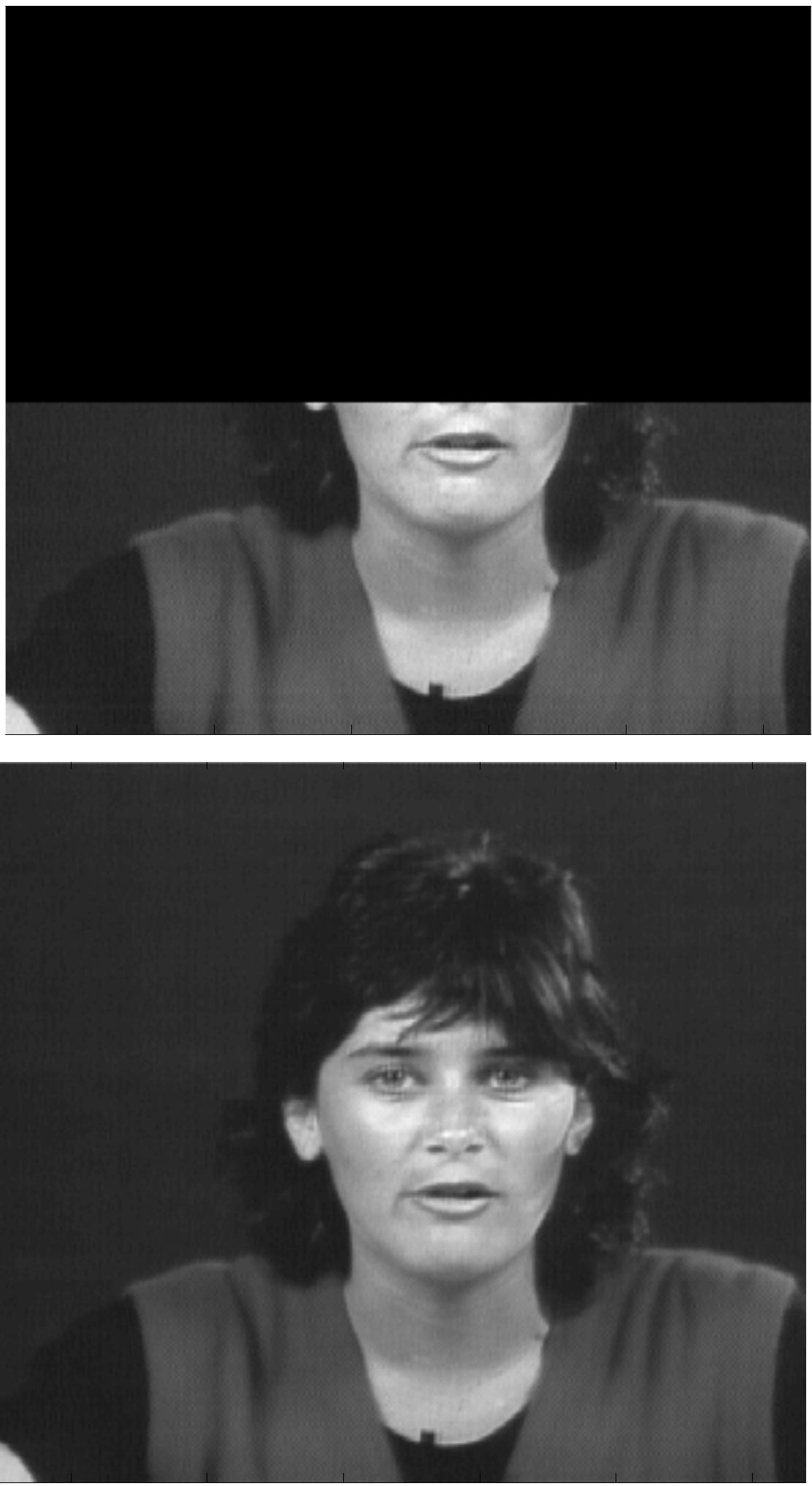}
	\label{fig:missa4}
}
\subfigure[]{
	\includegraphics[width=0.23\textwidth]{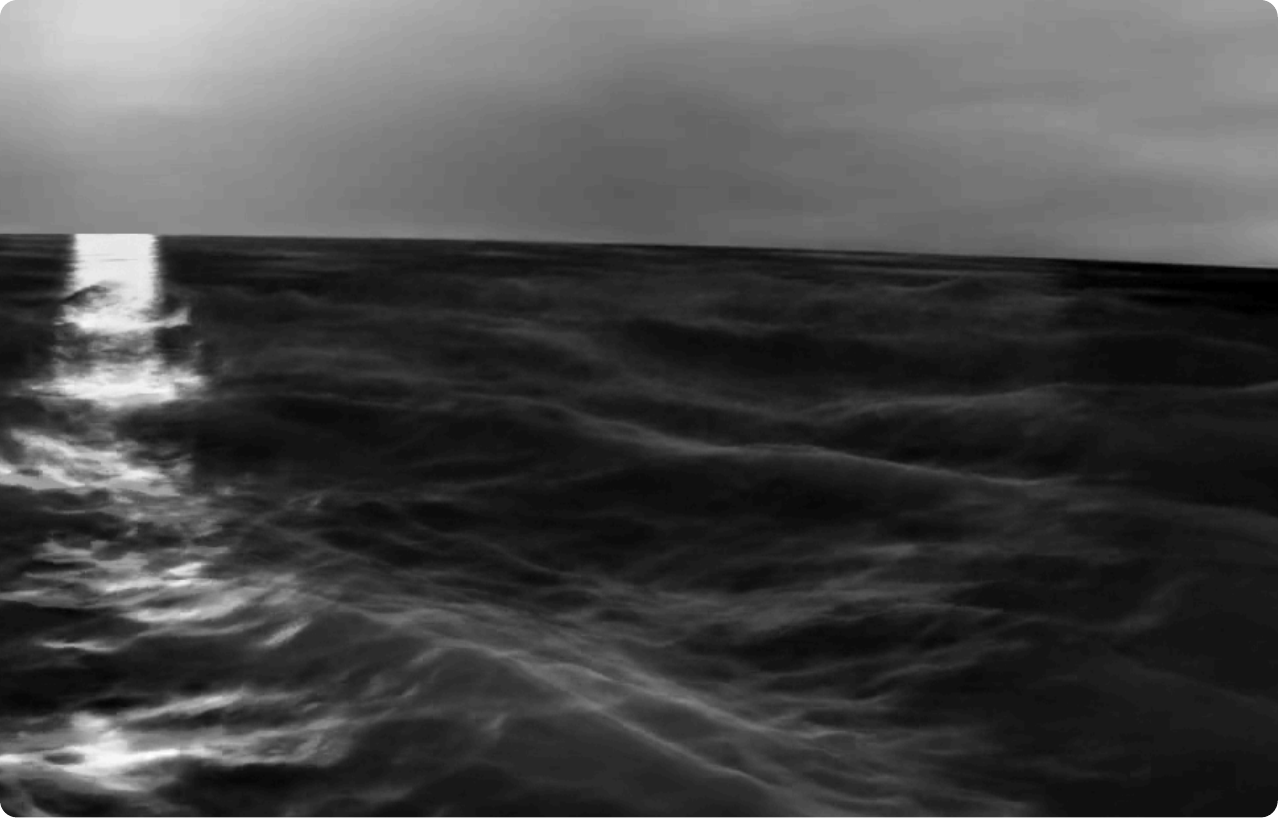}
	\label{fig:ocean1}
}
\subfigure[]{
	\includegraphics[width=0.23\textwidth]{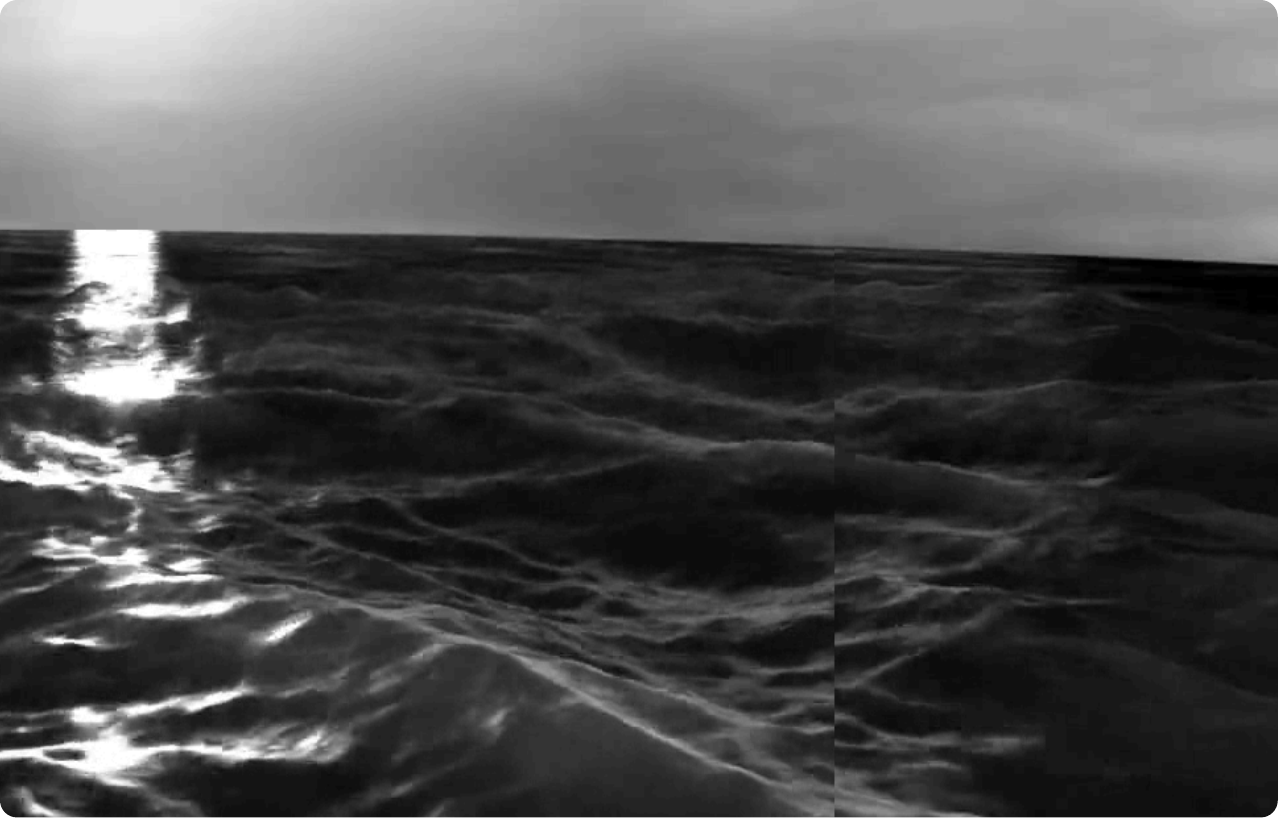}
	\label{fig:ocean2}
}
\subfigure[]{
	\includegraphics[width=0.23\textwidth]{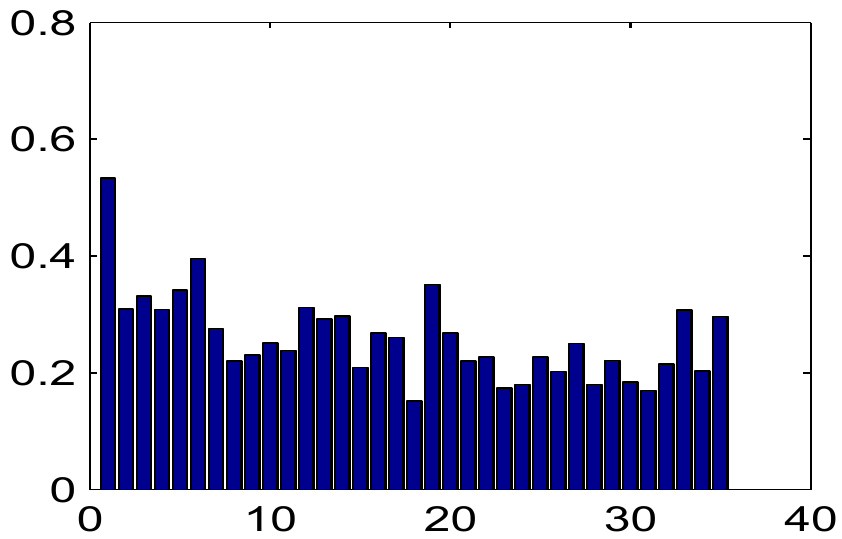}
	\label{fig:scalesDogInit}
}
\subfigure[]{
	\includegraphics[width=0.23\textwidth]{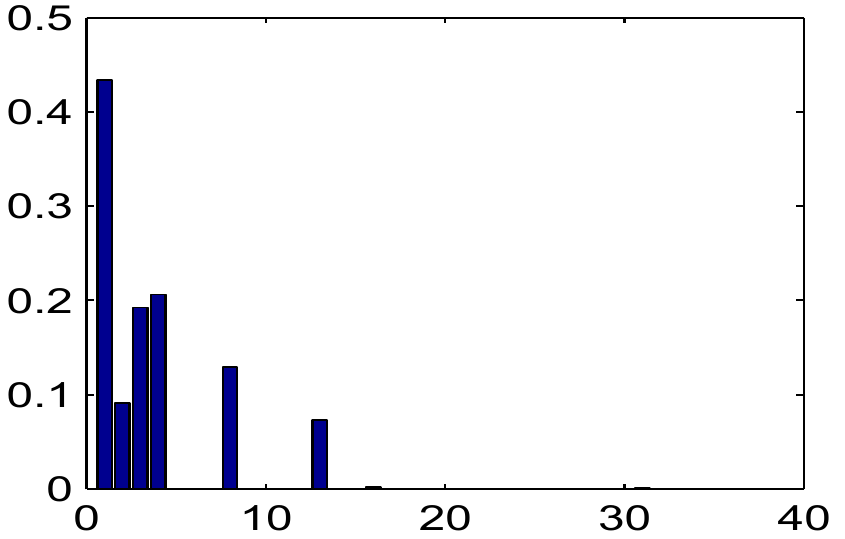}
	\label{fig:scalesDogOpt}
}
\end{center}
\caption{\small{
\subref{fig:missa1} and \subref{fig:missa3} demonstrate the reconstruction achieved by VGPDS and NN respectively for the most challenging frame \subref{fig:missa2} of the `missa' video, i.e.\ when translation occurs. \subref{fig:missa4} shows another example of the reconstruction achieved by VGPDS given the partially observed image. \subref{fig:ocean1} (VGPDS) and \subref{fig:ocean2} (NN) depict the reconstruction achieved for a frame of the `ocean' dataset. 
Finally, we demonstrate the ability of the model to automatically select the latent dimensionality by showing the initial lengthscales (fig: \subref{fig:scalesDogInit}) of the ARD kernel and the values obtained after training (fig: \subref{fig:scalesDogOpt}) on the `dog' data set.
}
}
\label{fig:video1}
\end{figure}
As can be seen in figure \ref{fig:video1}, VGPDS predicts pixels which are smoothly connected with the observed image, whereas the NN method cannot fit the predicted pixels in the overall context.

\par As a second task, we used our generative model to create new
samples and generate a new video sequence. This is most effective for
the `dog' video as the training examples were approximately periodic
in nature. The model was trained on 60 frames (time-stamps $[t_1,
t_{60}]$) and we generated the new frames which correspond to the next
40 time points in the future. The only input given for this generation
of future frames was the time stamp vector, $[t_{61}, t_{100}]$. The
results show a smooth transition from training to test and amongst the
test video frames. The resulting video of the dog continuing to run is
sharp and high quality. This experiment demonstrates the ability of
the model to reconstruct massively high dimensional images without
blurring. Frames from the result are shown in figure
\ref{fig:dog}. The full video is available in the supplementary
material.

\begin{figure}[ht]
\begin{center}
\subfigure[]{
	\includegraphics[width=0.23\textwidth]{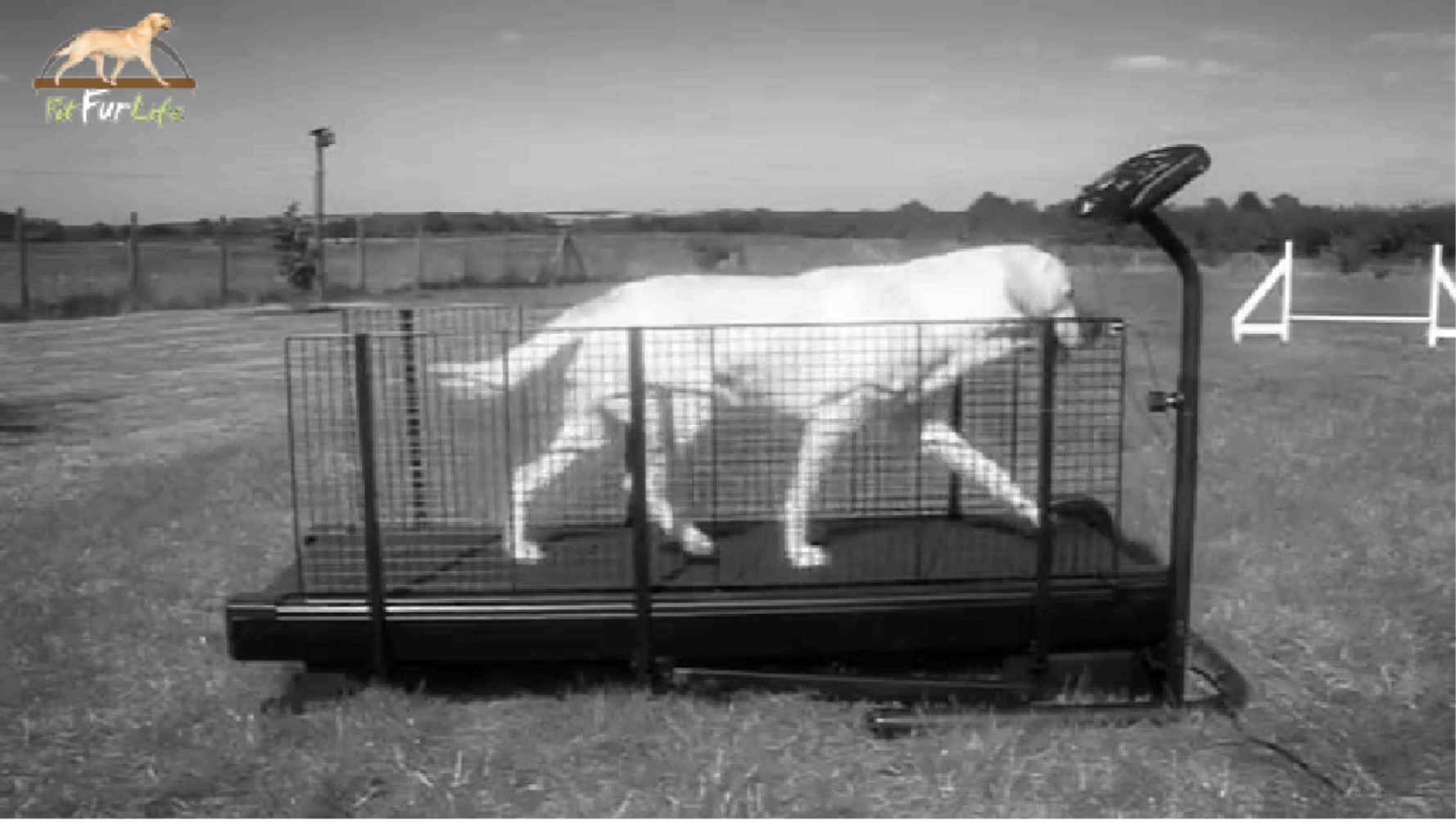}
	\label{fig:dogTrain}
}
\subfigure[]{
	\includegraphics[width=0.23\textwidth]{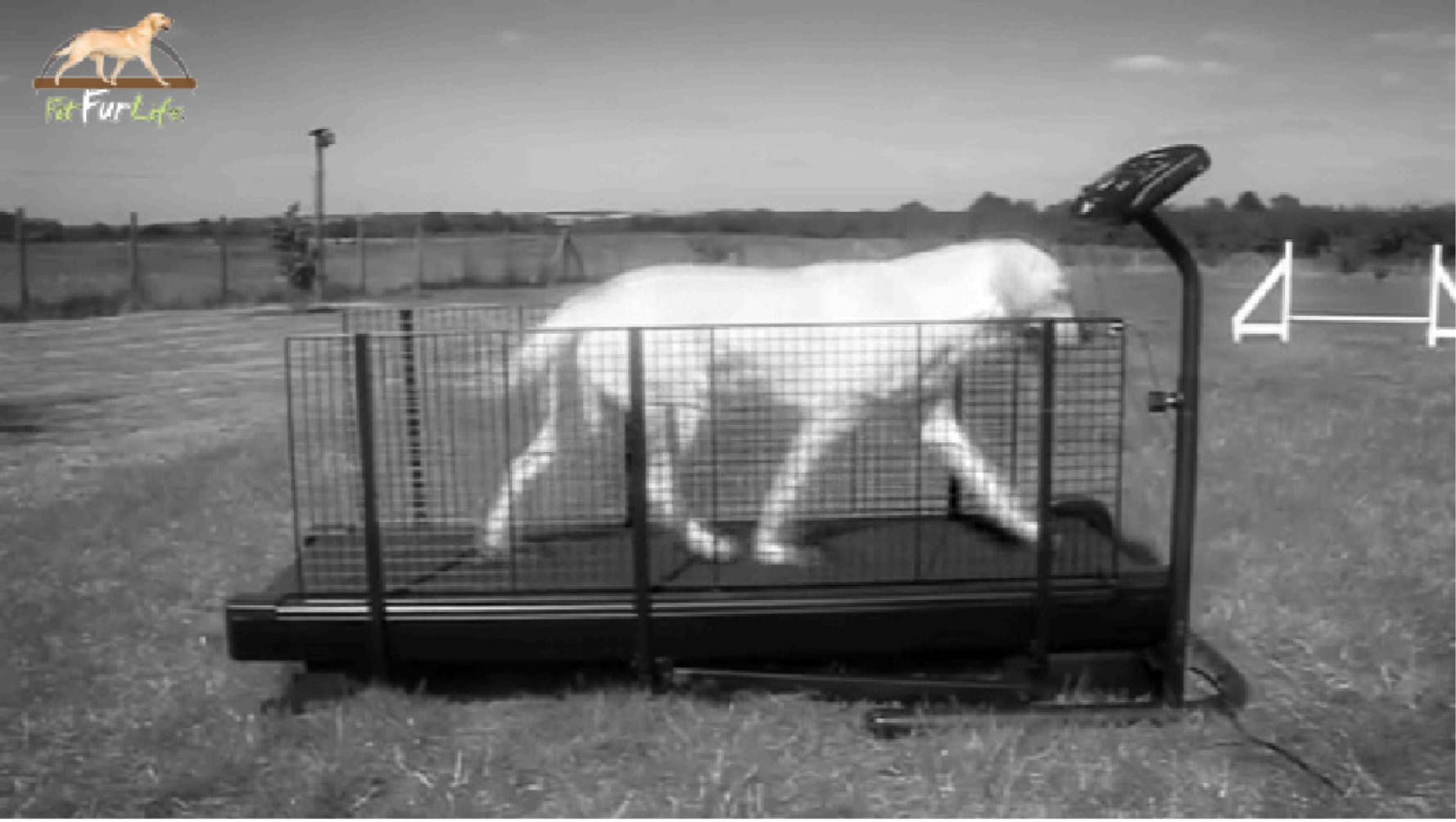}
	\label{fig:dogTest1}
}
\subfigure[]{
	\includegraphics[width=0.23\textwidth]{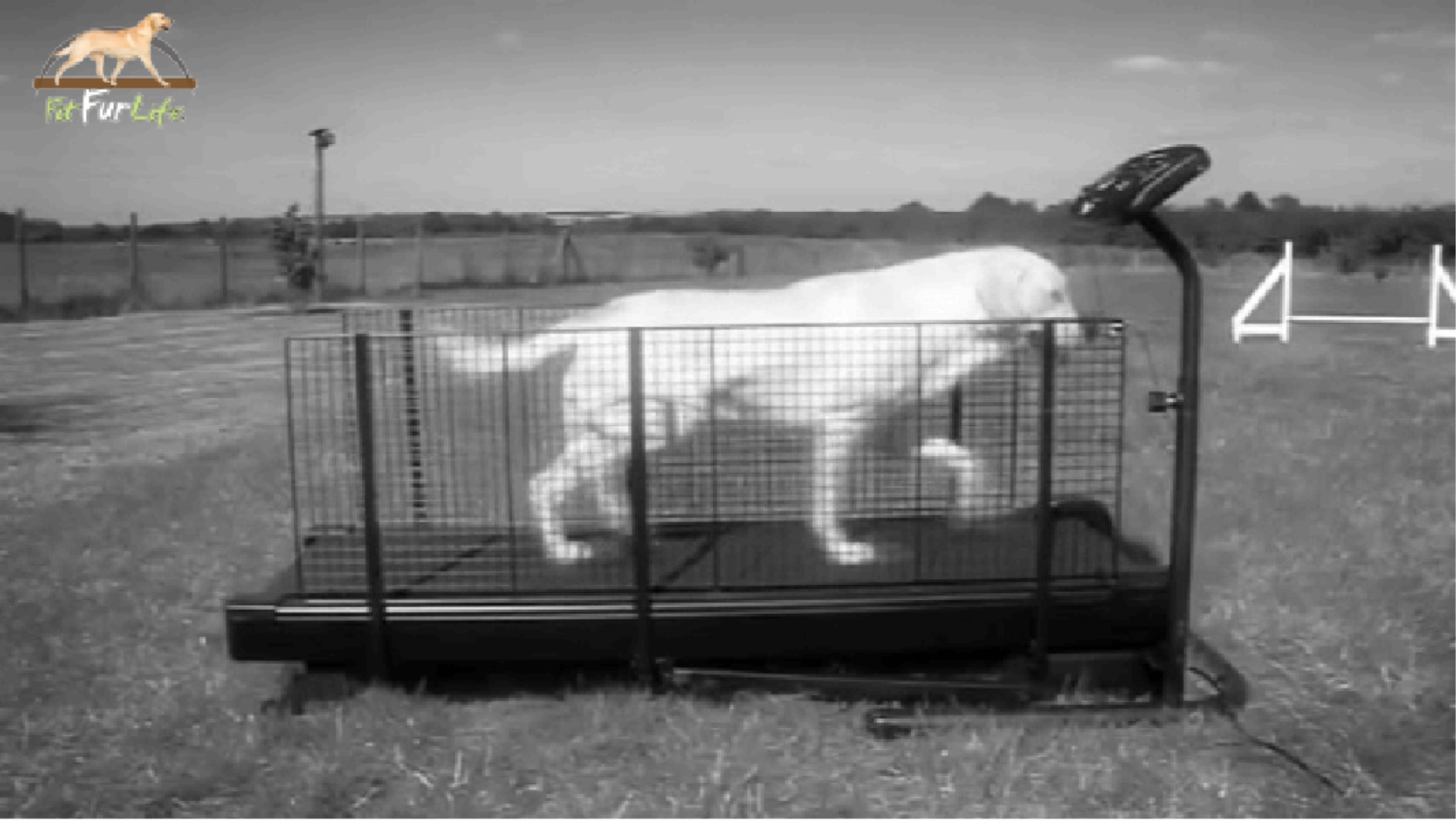}
	\label{fig:dogTest2}
}
\end{center}
\caption{ \small{
The last frame of the training video \subref{fig:dogTrain} is smoothly followed by the first frame \subref{fig:dogTest1} of the generated video. A subsequent generated frame can be seen in \subref{fig:dogTest2}}.}
\label{fig:dog}
\end{figure}


\section{Discussion and Future Work}

We have introduced a fully Bayesian approach for modeling dynamical
systems through probabilistic nonlinear dimensionality
reduction. Marginalizing the latent space and reconstructing data
using Gaussian processes results in a very generic model for capturing
complex, non-linear correlations even in very high dimensional data,
without having to perform any data preprocessing or exhaustive search
for defining the model's structure and parameters.


Our method's effectiveness has been demonstrated in two tasks;
firstly, in modeling human motion capture data and, secondly, in
reconstructing and generating raw, very high dimensional video
sequences. A promising future direction to follow would be to enhance
our formulation with domain-specific knowledge encoded, for example,
in more sophisticated covariance functions or in the way that data are
being preprocessed. Thus, we can obtain application-oriented methods
to be used for tasks in areas such as robotics, computer vision and
finance.

\subsubsection*{Acknowledgments}
Research was partially supported by the University of Sheffield Moody endowment fund and the Greek State Scholarships Foundation (IKY).
We also thank Colin Litster and ``Fit Fur Life'' for allowing us to use their video files as datasets.

\bibliographystyle{ieeetr}
\renewcommand*{\refname}{\begin{normalsize}References\end{normalsize}}
\bibliography{VGPDSarxiv}

\newpage
 \begin{center}
 \begin{Large}
 \textbf{
 Appendix
 } \\
 \end{Large}
 \end{center}
\appendix
\section{Derivation of the variational bound}

We wish to approximate the marginal likelihood:
\begin{equation}
\label{marginalLikelihoodSuppl}
p(Y | \bft) =  \int p( Y , F, X| \bft) \intd  X \intd F,
\end{equation}
by computing a lower bound:
\begin{align}
\mathcal{F}_v(q, \boldsymbol \theta) = {}& \int q(\mathit{\Theta}) \log 
		\frac{ p(Y , F , \mathit{X} | \mathbf{t})}
			 {q(\mathit{\Theta})}  \intd  X \intd F,
		 \label{jensens1Suppl}
\end{align}
This can be achieved by first augmenting the joint probability density of our model with inducing inputs $\tilde{X}$ along with their corresponding function values $U$:
\begin{equation}
 \label{augmentedJointSuppl}
p(Y,F, U,X,\tilde{X} | \bft) = \prod_{d=1}^D p(\mathbf{y}_d | \mathbf{f}_d) p(\mathbf{f}_d | \mathbf{u}_d, \mathit{X})
p(\bfu_d | \tilde{X})  p(X | \mathbf{t})
\end{equation}
where $p(\bfu_d | \tilde{X}) = \prod_{d=1}^D \mathcal{N} \left( \bfu_d | \mathbf{0}, K_{MM} \right)$ . For simplicity, $\tilde{X}$ is dropped from our
expressions for the rest of this supplementary material. Note that after including the inducing points, $p(\bff_d | \bfu_d, X)$
remains analytically tractable and it turns out to be \cite{rasmussen-williams}):
\begin{equation}
 \label{priorF2Suppl}
p(\bff_d | \bfu_d, X) =  \mathcal{N}  \left( \bff_d | K_{NM} K_{MM}^{-1} \bfu_d , K_{NN} - K_{NM} K_{MM}^{-1} K_{MN} \right).
\end{equation}
We are now able to define a variational distribution $q(\Theta)$ which factorises as:
For tractability we now define a variational density, $q(\Theta)$:
\begin{equation}
\label{varDistrSuppl}
q(\mathit{\Theta}) = q(F, U,X) = q(F | U, X) q(U) q(X) = \prod_{d=1}^D p(\bff_d | \bfu_d, X )q(\bfu_d) q(X),
\end{equation}
where $q(X) = \prod_{q=1}^Q \mathcal{N} \left( \bfx_q | \bfmu_q, S_q \right)$. 
Now, we return to \eqref{jensens1Suppl} and replace the joint distribution with its augmented version \eqref{augmentedJointSuppl} and the variational distribution with its factorised version \eqref{varDistrSuppl}:
\begin{align}
\mathcal{F}_v(q, \boldsymbol \theta) = {}& \int q(\mathit{\Theta}) \log 
		\frac{ p(Y,F, U,X | \bft)}
			 {q(F, U,X)}  \intd  X \intd F,
 	    \nonumber \\
= {}& \int \prod_{d=1}^D p(\bff_d | \bfu_d, X )q(\bfu_d) q(X) 
	    \log  \frac{\prod_{d=1}^D p(\mathbf{y}_d | \mathbf{f}_d) \cancel{p(\mathbf{f}_d | \mathbf{u}_d, \mathit{X})}
						p(\bfu_d | \tilde{X})  p(X | \mathbf{t})}
 	      		   {\prod_{d=1}^D \cancel{p(\bff_d | \bfu_d, X )}q(\bfu_d) q(X))}   \intd  X \intd F \nonumber \\
= {}& \int \prod_{d=1}^D p(\bff_d | \bfu_d, X )q(\bfu_d) q(X) 
		\log  \frac{\prod_{d=1}^D p(\mathbf{y}_d | \mathbf{f}_d) p(\bfu_d | \tilde{X})}
				   {\prod_{d=1}^D q(\bfu_d) q(X))}   \intd  X \intd F, \nonumber \\
- {}& \int \prod_{d=1}^D  q(X)   \log \frac{q(X)}{p(X | \mathbf{t})}   \intd  X \nonumber \\
= {}& \hat{\mathcal{F}}_v - \text{KL}(q \parallel p), \label{jensensSuppl}
\end{align}
with $\hat{\mathcal{F}}_v =\int q(X) \log p( Y | F ) p( F | X) \,
\mathrm{d} X \intd F = \sum_{d=1}^D \hat{\mathcal{F}}_d$. Both terms in \eqref{jensensSuppl} are analytically tractable, with the first having the same analytical solution as the one derived in \cite{BayesianGPLVM}. Further calculations in the the $\hat{\mathcal{F}}_v$ term reveal that the optimal setting for $q(\bfu_d)$ is also a Gaussian. More specifically, 
we have:
\begin{align}
\hat{\mathcal{F}}_v={}& \int q(\bfu_d) \log \frac{e^{\la \log N \left( \bfy_d | \bfa_d, \beta^{-1} I_d \right) \ra_{q(X)}}
		p(\bfu_d)}{q(\bfu_d)} d\bfu_d - A \label{boundFAnalytically5}
\end{align}
where $A$ is a collection of remaining terms and $\bfa_d$ is the mean of \eqref{priorF2Suppl}.
\eqref{boundFAnalytically5} is a KL-like quantity and, therefore, $q(\bfu_d)$ is optimally set to be the quantity appearing in the numerator of the above equation. So:
$$
q(\bfu_d) = e^{\la \log \mathcal{N} \left( \bfy_d | \bfa_d, \beta^{-1} I_d \right) \ra_{q(X)}}
		p(\bfu_d) ,
$$
exactly as in \cite{BayesianGPLVM}. This is a Gaussian distribution since $p(\bfu_d ) = \mathcal{N} (\bfu_d | \mathbf{0}, K_{MM} )$.

\par
The complete form of the jensen's lower bound turns out to be:
\begin{align}
\mathcal{F}_v(q, \boldsymbol \theta) = {}& \sum_{d=1}^{D} 
	\hat{\mathcal{F}}_d(q, \boldsymbol \theta) -  \text{KL}(q \parallel p) \nonumber \\
	= {}& 
	\sum_{d=1}^{D} 
		\log \left( 
		\frac{(\beta)^{\frac{N}{2}} \vert \mathit{K_{MM}} \vert ^\frac{1}{2} }
			 {(2\pi)^{\frac{N}{2}} \vert \beta \Psi_2 + \mathit{K_{MM}}  \vert ^\frac{1}{2} } 	
		 e^{-\frac{1}{2} \mathbf{y}^{T}_{d} W \mathbf{y}_d} 
		 \right) -
		 \frac{\beta \psi_0}{2} + \frac{\beta}{2} 
		 \text{Tr} \left( \mathit{K_{MM}^{-1}} \Psi_2 \right)  \nonumber \\
{}&	- \frac{Q}{2} \log \vert \mathit{K_t} \vert - \frac{1}{2} \sum_{q=1}^{Q}
	  \left[ \text{Tr} \left( \mathit{K_t}^{-1} \mathit{S_q} \right)	  
	  	   + \text{Tr} \left( \mathit{K_t}^{-1} \boldsymbol \mu_q \boldsymbol \mu_q^T \right) \right] 
	 + \frac{1}{2} \sum_{q=1}^Q \log \vert \mathit{S_q} \vert + const  \label{boundFinal}
\end{align}
where the last line corresponds to the KL term. Also:
\begin{equation}
\label{psis}
\Psi_0 = \text{Tr}(\langle \mathit{K_{NN}} \rangle_{q(\mathit{X})}) \;, \;\;
\Psi_1 = \langle \mathit{K_{NM}} \rangle_{q(\mathit{X})} \;, \;\;
\Psi_2 = \langle \mathit{K_{MN}} \mathit{K_{NM}} \rangle_{q(\mathit{X})}
\end{equation}
The $\Psi$ quantities can be computed analytically as in \cite{BayesianGPLVM}.


\section{Derivatives of the variational bound}
Before giving the expressions for the derivatives of the variational bound \eqref{jensensSuppl},
it should be reminded that the variational parameters $\mu_q$ and $S_q$ (for all $q$s) have been
reparametrised as $S_q = \left( \mathit{K}_t^{-1} + diag(\boldsymbol \lambda_q) \right)^{-1}  \text{ and }   \boldsymbol \mu_q = K_t \bar{\boldsymbol \mu}_q$, where the function $diag(\cdot)$ transforms a vector into a square diagonal matrix and vice versa. Given the above, the set of the parameters to be optimised is 
$( \boldsymbol \theta_f, \boldsymbol \theta_x, \{ \bar{\bfmu}_q, \boldsymbol \lambda_q \}_{q=1}^Q, \tilde{X}$. The gradient w.r.t the inducing points $\tilde{X}$, however, has exactly the same form as for $\boldsymbol \theta_f$ and, therefore, is not presented here. Also notice that from now on we will often use the term ``variational parameters'' to refer to the new quantities $\bar{\bfmu}_q$ and $\boldsymbol \lambda_q$. 

\textbf{Some more notation:} 
\begin{enumerate}
\item $\lambda_q$ is a scalar, an element of the vector $\boldsymbol \lambda_q$ which, in turn, is the main diagonal of the diagonal matrix $\Lambda_q$. 
\item $S_{ij} \triangleq S_{q;ij}$ the element of $S_q$ found in the $i$-th row and $j$-th column.
\item $\mathbf{s}_q \triangleq \lbrace S_{q;ii} \rbrace_{i=1}^N$, i.e. it is a vector with the diagonal of $S_q$.
\end{enumerate}

\subsection{Derivatives w.r.t the variational parameters}
\begin{equation}
    \label{derivVarParamSuppl}
\frac{\vartheta \mathcal{F}_v}{\vartheta \bar{\boldsymbol \mu}_q} 
=  K_t \left( \frac{\vartheta \hat{\mathcal{F}}}{\vartheta \boldsymbol \mu_q} - \bar{\boldsymbol \mu}_q \right)
\text{ and }
 \frac{\vartheta \mathcal{F}_v}{\vartheta \boldsymbol \lambda_q}
= - ( S_q \circ S_q) \left( \frac{\vv \hat{\mathcal{F}}}{\vv \mathbf{s}_q} + \frac{1}{2} \boldsymbol \lambda_q \right).
\end{equation}

where:

\begin{align}
 \frac{\hat{\mathcal{F}}(q, \boldsymbol \theta)}{\vartheta \mu_q}
{}& = - \frac{\beta D}{2} \frac{\vartheta \Psi_0}{\vartheta \mu_q}
    + \beta \text{Tr} \left(\frac{\vartheta \Psi_1^T}{\vartheta \mu_q} Y Y^T \Psi_1 A^{-1} \right) \nonumber \\
{}& + \frac{\beta}{2} \text{Tr} \left[ \frac{\vartheta \Psi_2}{\vartheta \mu_q}
       \left(
	  D K_{MM}^{-1} - \beta^{-1} D A^{-1} - A^{-1} \Psi_1^T Y Y^T \Psi_1 A^{-1}
       \right) \right] \label{derivFTildeEfficientComputationMu}
\end{align}

\begin{align}
 \frac{\vv \hat{\mathcal{F}}(q, \boldsymbol \theta)}{\vartheta S_{q;i,j}}
{}& = - \frac{\beta D}{2} \frac{\vartheta \Psi_0}{\vartheta S_{q;i,j}}
    + \beta \text{Tr} \left(\frac{\vartheta \Psi_1^T}{\vartheta S_{q;i,j}} Y Y^T \Psi_1 A^{-1} \right) \nonumber \\
{}& + \frac{\beta}{2} \text{Tr} \left[ \frac{\vartheta \Psi_2}{\vartheta S_{q;i,j}}
       \left(
	  D K_{MM}^{-1} - \beta^{-1} D A^{-1} - A^{-1} \Psi_1^T Y Y^T \Psi_1 A^{-1}
       \right) \right] \label{derivFTildeEfficientComputationS}
\end{align}

with $A=\beta^{-1}K_{MM}+\Psi_2$.


\subsection{Derivatives w.r.t $\boldsymbol \theta = (\boldsymbol \theta_f, \boldsymbol \theta_x)$ and $\beta$}
Given that the KL term involves only the temporal prior, its gradient w.r.t the parameters $\boldsymbol \theta_f$ is zero. Therefore:
\begin{equation}
   \label{DerivativeOfFComplete}
      \frac{\vartheta \mathcal{F}_v}{\vartheta \theta_f} = \frac{\vartheta \hat{\mathcal{F}}}{\vartheta \theta_f}
\end{equation}

  with:

\begin{align}
\frac{\vartheta \hat{\mathcal{F}}}{\vartheta \theta_f} {}& = \text{const} - 
\frac{\beta D}{2} \frac{\vartheta \Psi_0}{\vartheta \theta_f}
 + \beta \text{Tr} \left(\frac{\vartheta \Psi_1^T}{\vartheta \theta_f} Y Y^T \Psi_1 A^{-1} \right) \nonumber \\
{}& + \frac{1}{2} \text{Tr} \left[ \frac{\vartheta K_{MM}}{\vartheta \theta_f}
        \left(
	   D K_{MM}^{-1} - \beta^{-1} D A^{-1} - A^{-1} \Psi_1^T Y Y^T \Psi_1 A^{-1} - \beta D K_{MM}^{-1} \Psi_2 K_{MM}^{-1} 
         \right) \right] \nonumber \\
{}& + \frac{\beta}{2} \text{Tr} \left[ \frac{\vartheta \Psi_2}{\vartheta \theta_f} \;\;\;\;
       \left(
	  D K_{MM}^{-1} - \beta^{-1} D A^{-1} - A^{-1} \Psi_1^T Y Y^T \Psi_1 A^{-1}
       \right) \right] \label{DerivativeOfFtildeComplete}
\end{align}

The expression above is identical for the derivatives w.r.t the inducing points.
For the gradients w.r.t the $\beta$ term, we have a similar expression:

\begin{align}
\frac{\vartheta \hat{\mathcal{F}}}{\vartheta \beta} ={}&
  \frac{1}{2} \Big[ 
      D \left( \text{Tr}(K_{MM}^{-1} \Psi_2) + (N-M)\beta^{-1} - \Psi_0 \right) - \text{Tr}(Y Y^\T)
	  + \text{Tr}(A^{-1}\Psi_1^\T Y Y^\T \Psi_1) \nonumber \\
   +{}& \beta^{-2} D \text{Tr} ( K_{MM} A^{-1} ) + \beta^{-1} \text{Tr} \left( K_{MM}^{-1} A^{-1} \Psi_1^\T Y Y^\T \Psi_1 A^{-1} \right) \Big]
\label{derivb2}
\end{align}

In contrast to the above, the term $\hat{\mathcal{F}}_v$ does involve parameters $\boldsymbol \theta_x$, because it involves the variational parameters that are now reparametrized with $K_t$, which in turn depends on $\boldsymbol \theta_x$. 
To demonstrate that, we will forget for a moment the reparametrization of $S_q$ and we will express the bound as $F(\boldsymbol \theta_x, \mu_q (\boldsymbol \theta_x))$ (where $\mu_q (\boldsymbol \theta_x) = K_t \bar{\boldsymbol \mu_q}$) so as to show explicitly the dependency on the variational mean which is now a function of $\boldsymbol \theta_x$. Our calculations must now take into account the term
$
\left( \frac{\vartheta \hat{\mathcal{F}}(\boldsymbol \mu_q)}{\vartheta \boldsymbol \mu_q} \right)^\T
       \frac{\vartheta \mu_q (\boldsymbol \theta_x)}{\vartheta \boldsymbol \theta_x}
$
that is what we ``miss'' when we consider $\mu_q(\boldsymbol \theta_x) = \boldsymbol \mu_q$:
\begin{align}
\frac{\vartheta \mathcal{F}_v(\boldsymbol \theta_x, \mu_q(\boldsymbol \theta_x))}{\vartheta \theta_x} = {}&
	\frac{\vartheta \mathcal{F}_v(\boldsymbol \theta_x, \boldsymbol \mu_q)}{\vartheta \theta_x} 
  +  \left( \frac{\vartheta \hat{\mathcal{F}}(\boldsymbol \mu_q)}{\vartheta \boldsymbol \mu_q} \right)^\T
            \frac{\vartheta \mu_q(\boldsymbol \theta_x)}{\vartheta \theta_x} \nonumber \\
= {}&
 \cancel{
    \frac{\vartheta \hat{\mathcal{F}}(\boldsymbol \mu_q)}{\vartheta \theta_x}
  } +
  \frac{\vv (-\text{KL})(\boldsymbol \theta_x, \boldsymbol \mu_q(\boldsymbol \theta_x))}{\vartheta \theta_x}
+  \left( \frac{\vartheta \hat{\mathcal{F}}(\boldsymbol \mu_q)}{\vartheta \boldsymbol \mu_q} \right)^\T
            \frac{\vartheta \mu_q(\boldsymbol \theta_x)}{\vartheta \theta_x}
\label{meanReparamDerivFTheta}
\end{align}

We do the same for $S_q$ and then we can take the resulting equations and replace $\bfmu_q$ and $S_q$ with their equals so as to take the final expression which only contains $\bar{\bfmu}_q$ and $\boldsymbol \lambda_q$:

\begin{align}
\frac{\vartheta \mathcal{F}_v(\boldsymbol \theta_x, \mu_q(\boldsymbol \theta_x), S_q(\boldsymbol \theta_x))}{\vartheta \theta_x}
={}& \text{Tr} \bigg[
\Big[ - \frac{1}{2} \left( \hat{B}_q K_t \hat{B}_q + \bar{\bfmu}_q \bar{\bfmu}_q^\T \right) \nonumber \\
+{}& \left( I - \hat{B}_q K_t \right)
 diag \left(  \frac{\vv \hat{\mathcal{F}}}{\vv \mathbf{s}_q} \right)
			 \left( I - \hat{B}_q K_t \right)^\T \Big]
			  \frac{\vv K_t}{\vv \theta_x} \bigg] 	\nonumber \\	
+{}&  \left( \frac{\vartheta \hat{\mathcal{F}}( \boldsymbol \mu_q)}{\vartheta \boldsymbol \mu_q} \right)^\T
					\frac{\vv K_t}{\vv \theta_x} \bar{\boldsymbol \mu}_q 
\label{CompleteBoundDerivThetatB}
\end{align}
where $\hat{B}_q = \Lambda_q^{\frac{1}{2}} \widetilde{B}_q^{-1} \Lambda_q^{\frac{1}{2}}$.
and $\tilde{B}_q = I + \Lambda_q^{\frac{1}{2}} K_t \Lambda_q^{\frac{1}{2}}$. Note that by using this
$\tilde{B}_q$ matrix (which has eigenvalues bounded below by one) we have an expression which, when implemented, leads to more numerically stable computations, as explained in \cite{rasmussen-williams} page 45-46.

\section{Predictions}

\subsection{Predictions given only the test time points \label{supplUnobservedData}}
To approximate the predictive density, we will need to introduce the underlying latent function values $F_* \in \mathbb{R}^{N_* \times D}$ (the noisy-free version of $Y_*$) and the latent variables $X_* \in \mathbb{R}^{N_* \times Q}$. We  write the predictive density as
\begin{eqnarray}
p(Y_* | Y) & = & \int p(Y_*, F_*, X_*| Y_*, Y)  \intd  F_* \intd  X_* =  \int p(Y_* | F_*)  p(F_*|X_*, Y) p(X_*|  Y) \intd  F_* \intd  X_* .
\label{eq:predictive1Suppl}
\end{eqnarray}
The term $p(F_* |X_*, Y)$ is approximated according to
\begin{eqnarray}
q(F_*|X_*) & = & \int \prod_{d \in D} p(\bff_{*,d} | \bfu_d, X_*)  q(\bfu_d) \intd  \bfu_d 
	    = \prod_{d \in D} q(\bff_{*,d} | X_*)  ,
\end{eqnarray}
where $q(\bff_{*,d} | X_*)$ is a Gaussian that can be computed analytically.
The term $p(X_*| Y)$ in eq. (\ref{eq:predictive1Suppl}) is approximated by
a Gaussian variational distribution $q(X_*)$,
\begin{align}
p(X_* | Y) \approx {}& \int  p(X_* | X) q(X) \intd  X = \la  p(X_* | X) \ra_{q(X)} = q(X_*) = \prod_{q=1}^Q q(\bfx_{*,q}),\label{qxstarSuppl}
\end{align}
where $p( X_{*,q} | X)$ can be found from the conditional GP prior
(see \cite{rasmussen-williams}). We can then write
\begin{equation}
\label{qxstar2Suppl}
\bfx_{*,q} = \boldsymbol \alpha \bfx_q + \boldsymbol \epsilon,
\end{equation} 
where $\boldsymbol \alpha = K_{*N}K_t^{-1}$ and 
$\boldsymbol \epsilon \sim \mathcal{N} \left( \bfz, K_{**} - K_{*N K_t^{-1} K_{N*}}\right)$. Also, $K_t = k_x(\bft, \bft)$, $K_{*N} = k_x(\bft_*, \bft)$ and $K_{**} = k_x(\bft_* \bft_*)$. 
Given the above, we know a priori that \eqref{qxstarSuppl} is a Gaussian and by taking expectations over $q(X)$ in the r.h.s. of \eqref{qxstar2Suppl} we find the mean and covariance of $q(X_*)$. Substituting for the equivalent forms of $\bfmu_q$ and $S_q$ from section \ref{optimisation} we obtain the final solution
\begin{align}
 \mu_{x_{*,q}} = {}& \bfk_{*N} \bar{\mu}_q \\
  \text{var}(x_{*,q}) = {}& k_{**} - \bfk_{*N} (K_t + \Lambda_q^{-1})^{-1} \bfk_{N*}.
\end{align}
\eqref{eq:predictive1Suppl} can then be written as:
\begin{align} 
p(Y_*| Y) {}& =  \int p(Y_*| F_*)  q(F_*|X_*) q(X_*) \intd  F_* \intd  X_* = \int p(Y_* | F_*) \la q(F_* | X_*) \ra_{q(X_*)} \intd  F_* \label{eq:predictive2Suppl}
\end{align}
Although the expectation appearing in the above integral is not a Gaussian, its moments can be found analytically \cite{rasmussen-williams, Girard03gaussianprocess},
\begin{align}
 \mathbb{E}(F_*) ={}&  B^\T \Psi_1^* \label{meanFstarSuppl} \\
 \text{Cov}(F_*) ={}& B^\T \left( \Psi_2^* - \Psi_1^* (\Psi_1^*)\T \right) B + \Psi_0^* I - \text{Tr} \left[ \left( K_{MM}^{-1} - \left( K_{MM} + \beta \Psi_2 \right)^{-1} \right) \Psi_2^* \right] I,
\end{align}
where $B = \beta \left( K_{MM} + \beta \Psi_2 \right)^{-1} \Psi_1^\T
Y$, $\Psi_0^* = \la k_f(X_*, X_*) \ra$, $\Psi_1^* = \la K_{M*} \ra$
and $\Psi_2^* = \la K_{M*} K_{*M} \ra$. All expectations are taken
w.r.t. $q(X_*)$ and can be calculated analytically, while $K_{M*}$
denotes the cross-covariance matrix between the training inducing
inputs $\tilde{X}$ and $X_*$. Finally, since $Y_*$ is just a noisy version of
$F_*$, the mean and covariance of \eqref{eq:predictive2Suppl} is just
computed as: $\mathbb{E}(Y_*) = \mathbb{E}(F_*)$ and $\text{Cov}(Y_*)
= \text{Cov}(F_*) + \beta^{-1} I_{N_*}$.

\subsection{Predictions given the test time points and partially observed outputs}

The expression for the predictive density $p(Y_*^m | Y_*^p, Y)$ follows exactly as in section \ref{supplUnobservedData} but we need to compute probabilities for $Y_*^m$ instead of $Y_*$ and $Y$ is replaced with $(Y, Y_*^p)$ in all conditioning sets. Similarly, $F$ is replaced with $F^m$. Now $q(X_*)$ cannot be found analytically as in section \ref{supplUnobservedData}; instead, it is optimised so that $Y_*^p$ are taken into account. 
This is done by maximising the variational lower bound on the marginal likelihood:
\begin{align}
p(Y_*^p, Y) ={}&  \int p(Y_*^p, Y|X_*, X) p(X_*, X) \intd  X_* \intd  X \nonumber \\
			={}&  \int p(Y^m | X) p(Y_*^p, Y^p|X_*, X) p(X_*, X) \intd  X_* \intd  X,  \nonumber
\end{align}  
Notice that here, unlike the main paper, we work with the likelihood after marginalising $F$, for simplicity.
Assuming a variational distribution 
$q(X_*, X)$ and using Jensen's inequality we obtain the 
lower bound 
\begin{eqnarray}
& & \int q(X_*, X) \log \frac{ p(Y^m | X) 
p(Y_*^p, Y^p|X_*, X) p(X_*,X)}{ q(X_*, X)} \intd  X_* \intd  X \nonumber \\ 
& = & \int q(X) \log p(Y^m | X) \intd  X 
+  \int q(X_*,X) \log p(Y_*^p, Y^p|X_*, X) \intd  X_* \intd  X  \nonumber \\
& - & \text{KL}[q(X_*,X) || p(X_*, X)] \label{partialPredLowerBoundSuppl}
\end{eqnarray}  
This quantity can now be maximized in the same manner as for the bound
of the training phase. Unfortunately, this means that the variational
parameters that are already optimised from the training procedure
cannot be used here because $X$ and $X_*$ are coupled in $q(X_*,X)$. A
much faster but less accurate method would be to decouple the test
from the training latent variables by imposing the factorisation
$q(X_*, X) = q(X) q(X_*)$. Then, equation
\eqref{partialPredLowerBoundSuppl} would break into terms containing $X$,
$X_*$ or both. The ones containing only $X$ could then be treated as
constants.

\section{Additional results from the experiments}
\begin{figure}[ht]
\begin{center}
\subfigure[]{
	\includegraphics[width=0.4\textwidth]{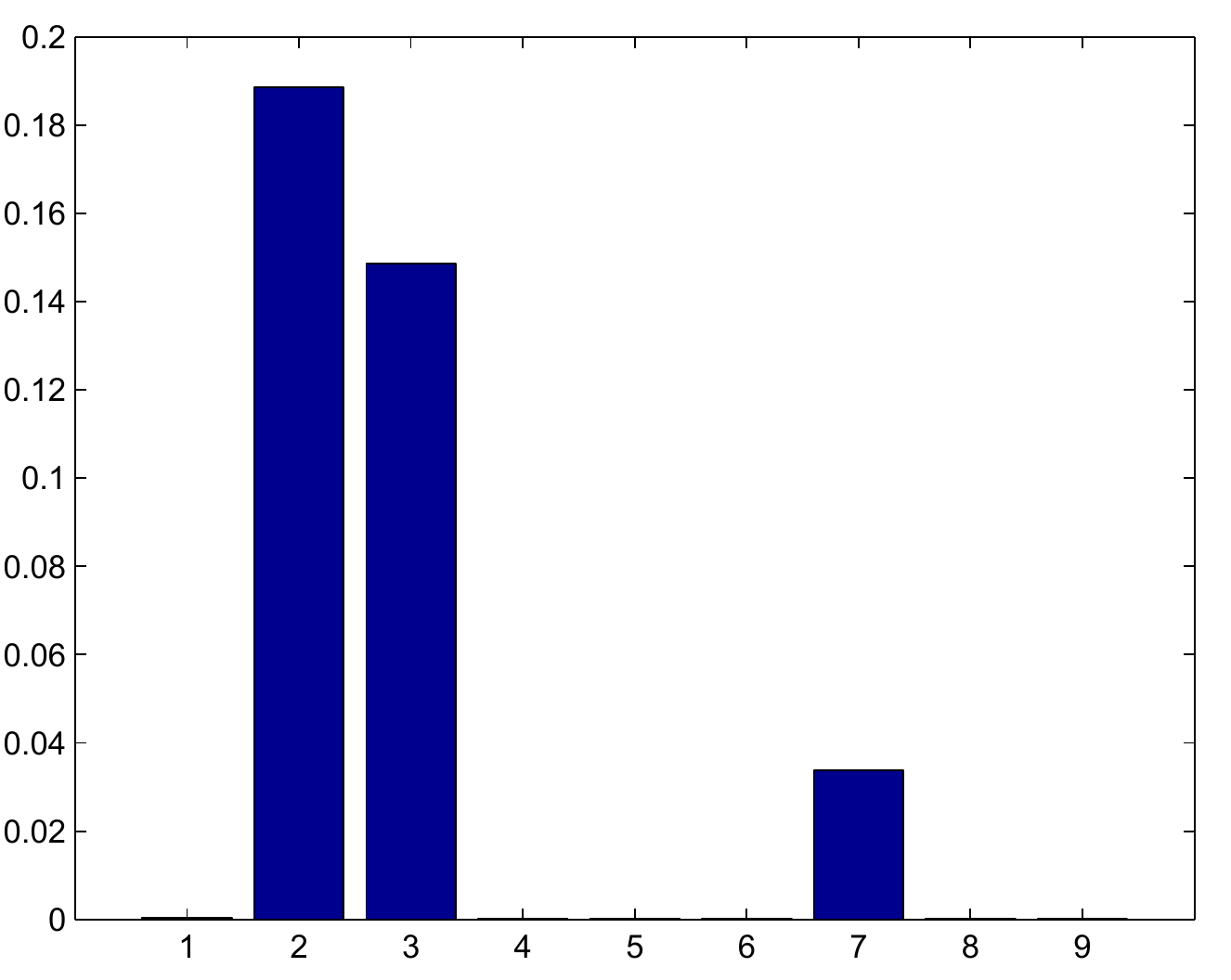}
	\label{fig:suppMocap1}
}
\subfigure[]{
	\includegraphics[width=0.4\textwidth]{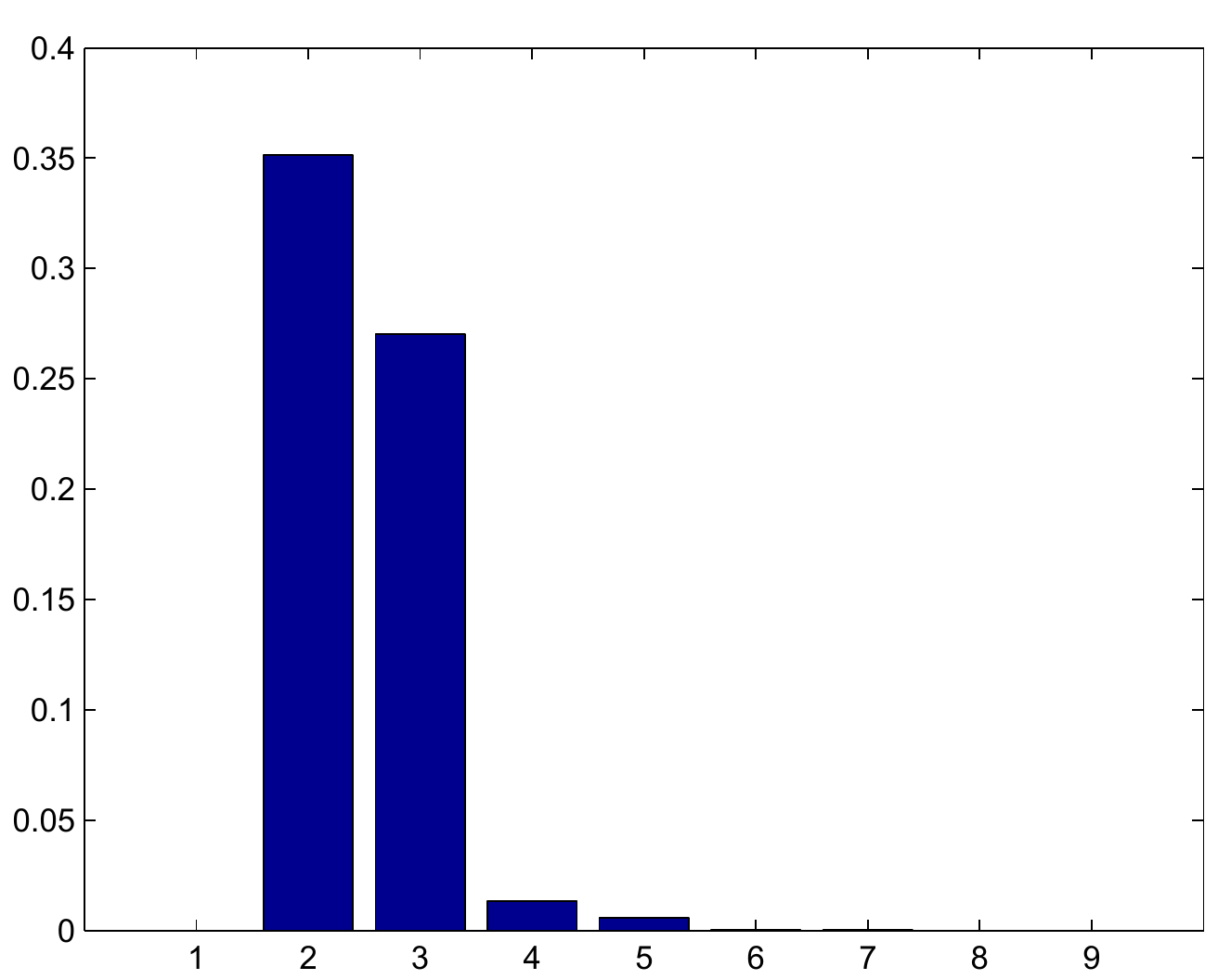}
	\label{fig:suppMocap2}
}
\end{center}
\caption{\small{
The values of the scales of the ARD kernel after training on the motion capture dataset using the RBF (fig: \subref{fig:suppMocap1}) and the Matern (fig: \subref{fig:suppMocap2}) kernel to model the dynamics for VGPDS. The scales that have zero value ``switch off'' the corresponding dimension of the latent space. The latent space is, therefore, 3-D for \subref{fig:suppMocap1} and 4-D for \subref{fig:suppMocap2}. Note that the scales were initialized with very similar values.
}
}
\label{fig:supplMocap1}
\end{figure}

\begin{figure}[ht]
\begin{center}
\subfigure[]{
	\includegraphics[width=0.48\textwidth]{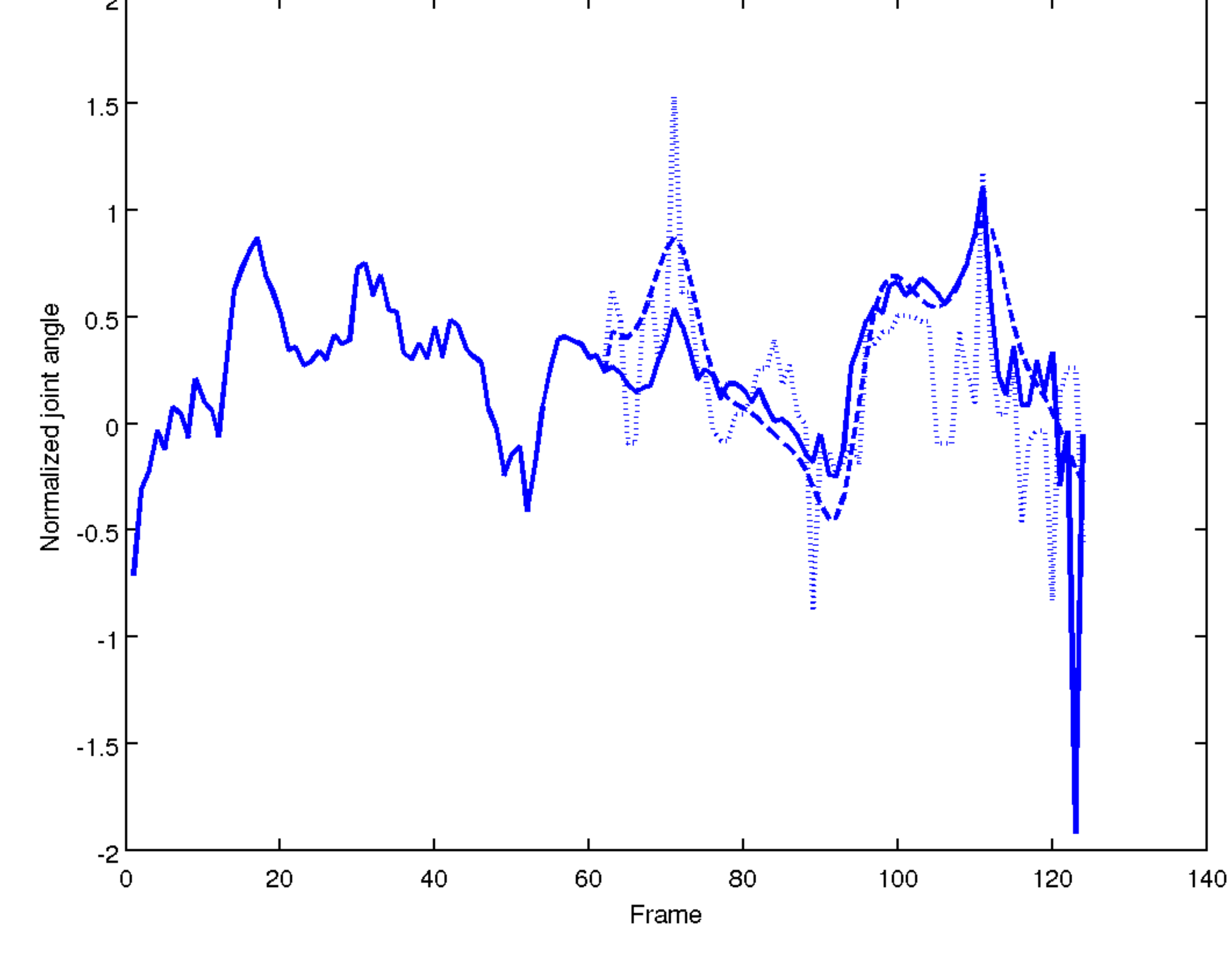}
	\label{fig:suppMocap3}
}
\subfigure[]{
	\includegraphics[width=0.48\textwidth]{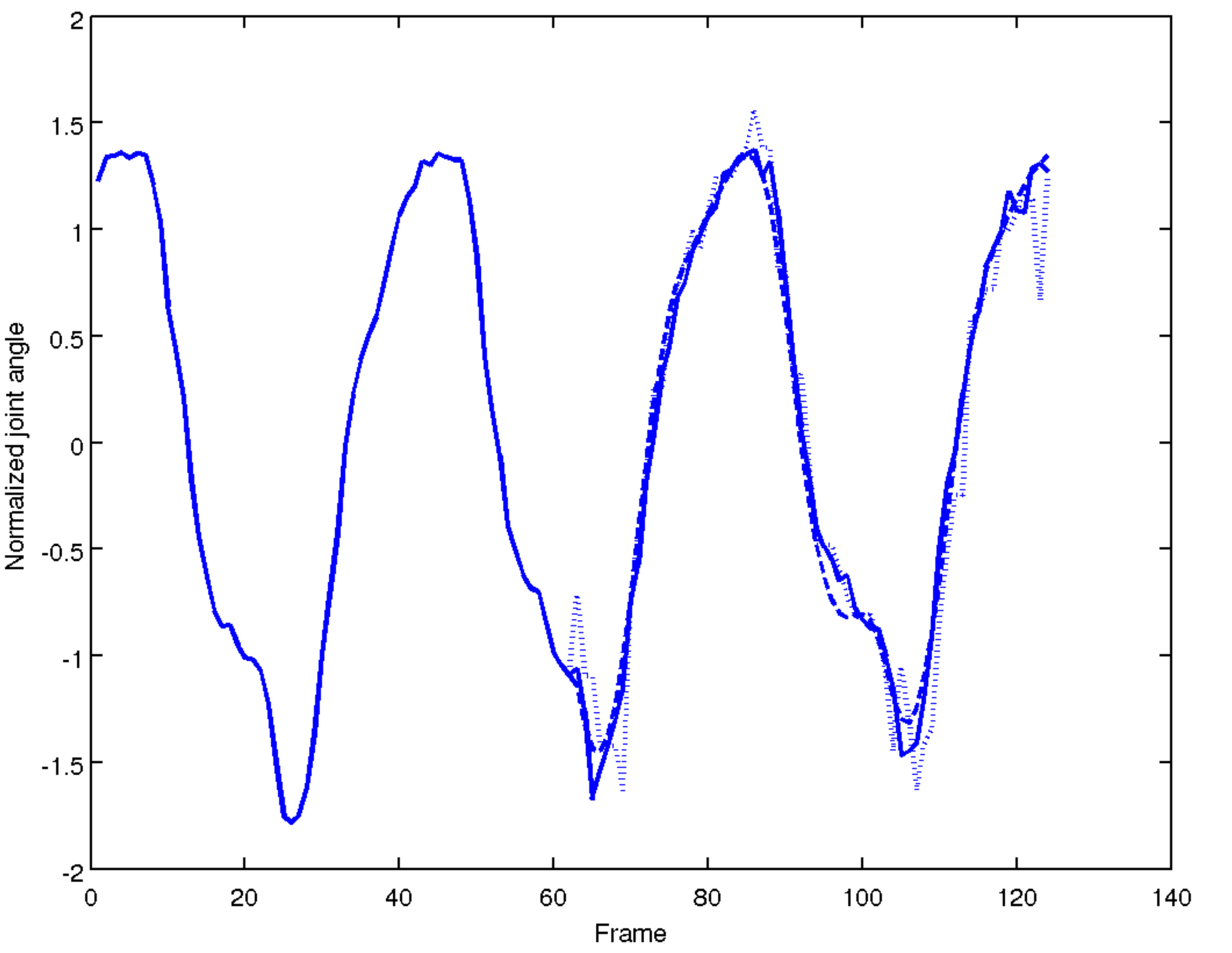}
	\label{fig:suppMocap4}
}
\end{center}
\caption{\small{
The prediction for two of the test angles for the body (fig: \ref{fig:suppMocap3}) and for the legs part (fig: \ref{fig:suppMocap3}). Continuous line is the original test data, dotted line is nearest neighour in scaled space, dashed line is VGPDS (using the RBF kernel for the body reconstruction and the Matern for the legs).
}
}
\label{fig:supplMocap2}
\end{figure}

\begin{figure}[ht]
\begin{center}
\subfigure[]{
	\includegraphics[width=0.23\textwidth]{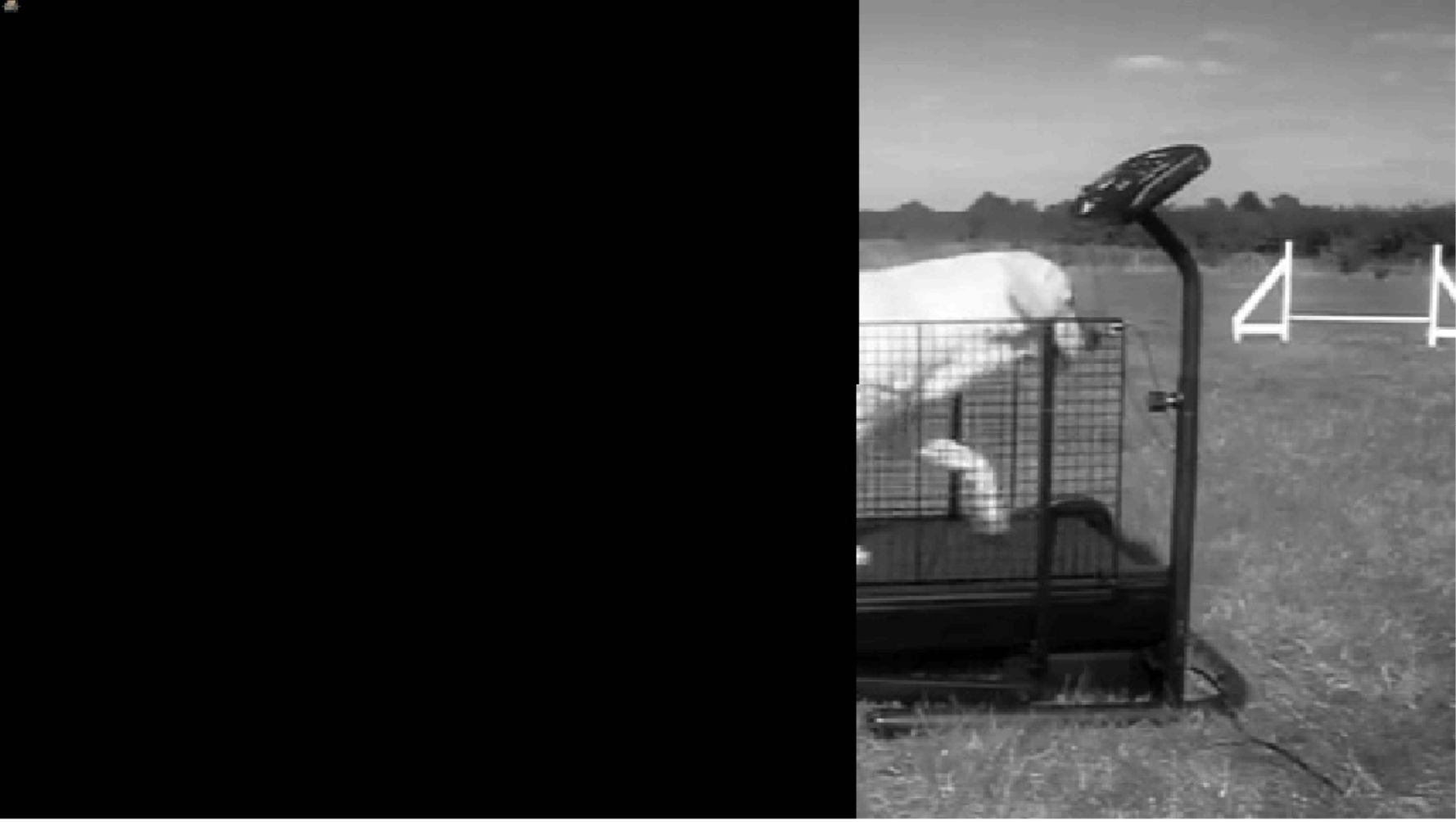}
	\label{fig:suppDog1}
}
\subfigure[]{
	\includegraphics[width=0.23\textwidth]{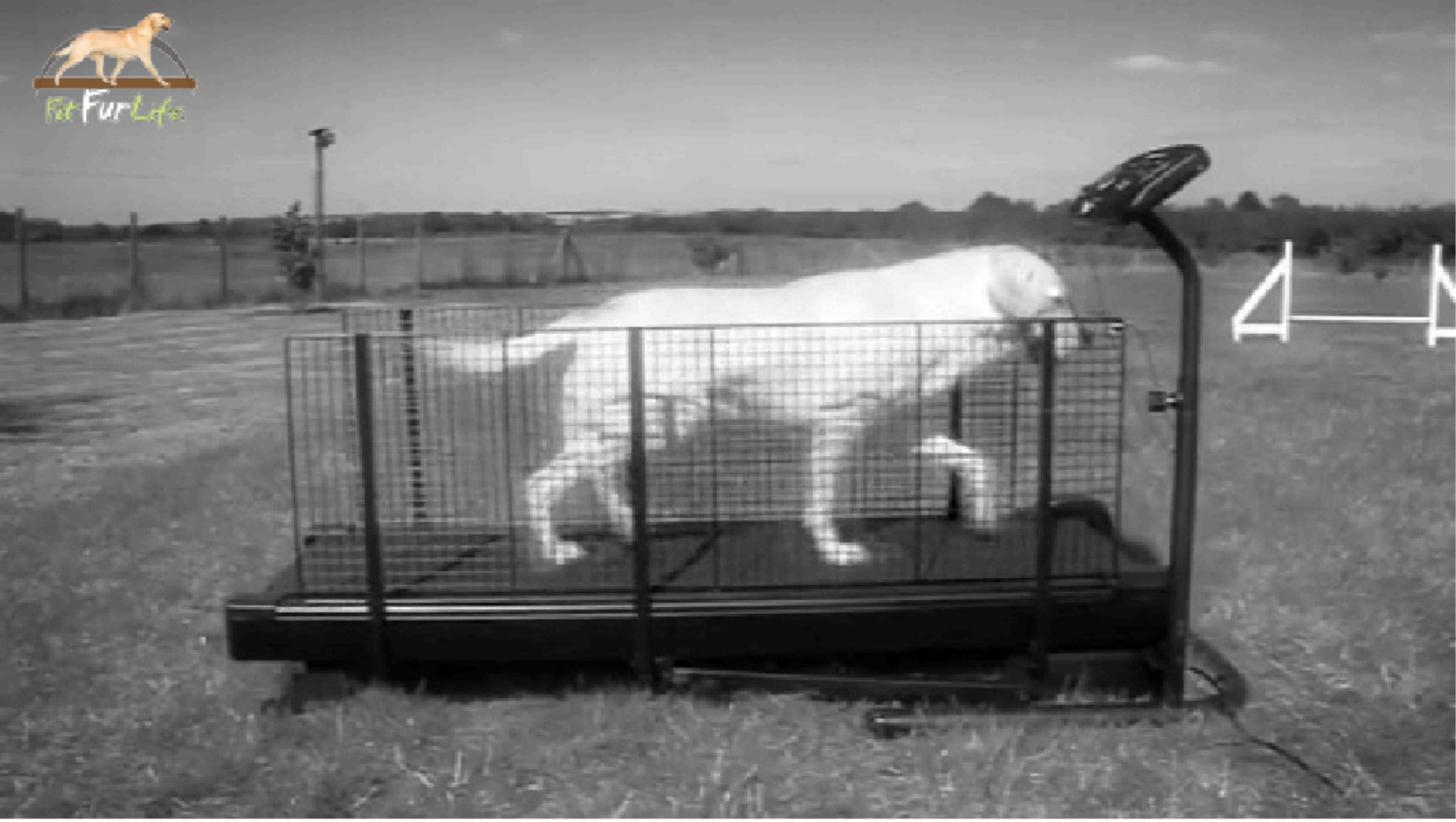}
	\label{fig:suppDog2}
}
\subfigure[]{
	\includegraphics[width=0.23\textwidth]{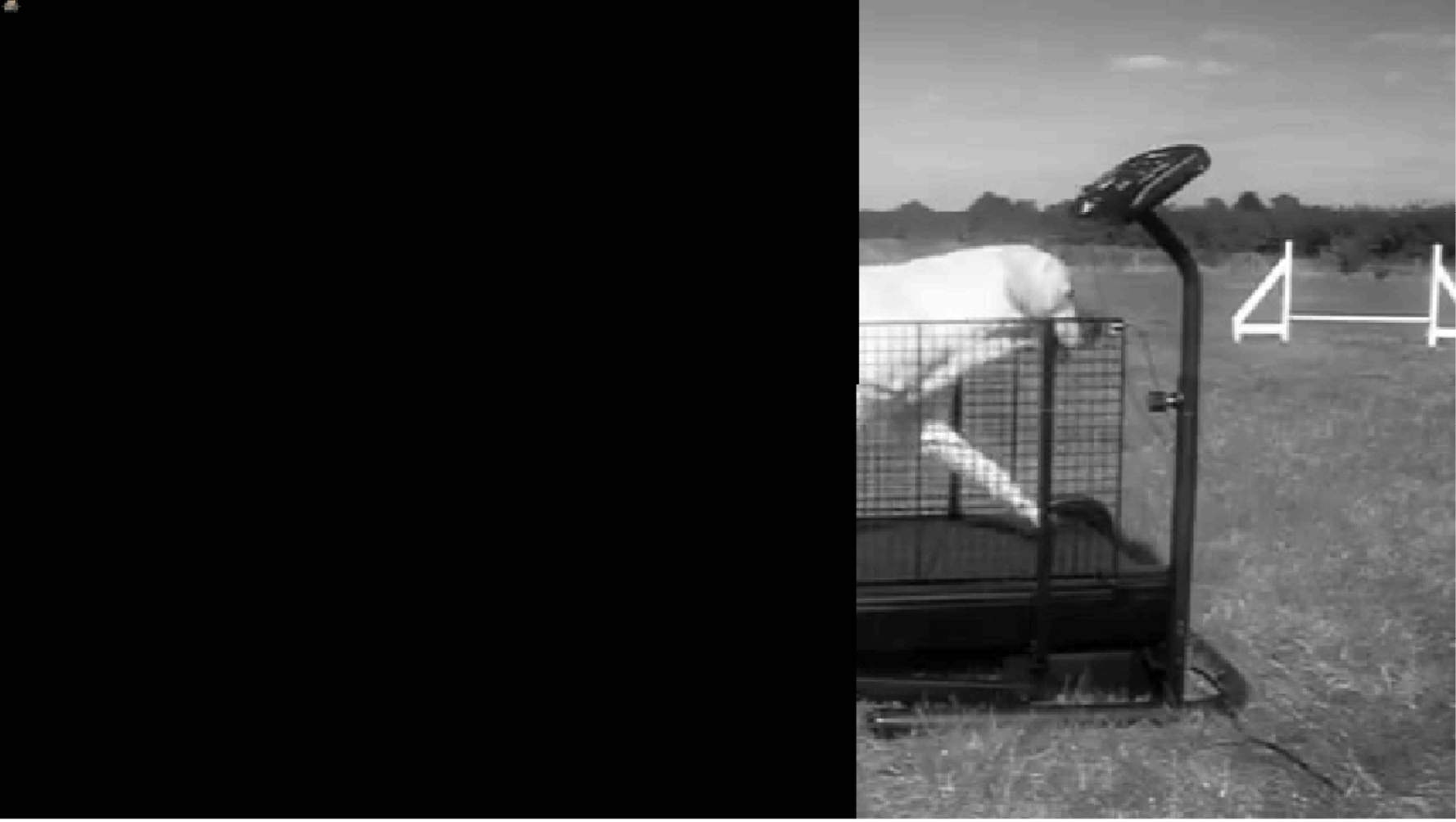}
	\label{fig:suppDog3}
}
\subfigure[]{
	\includegraphics[width=0.23\textwidth]{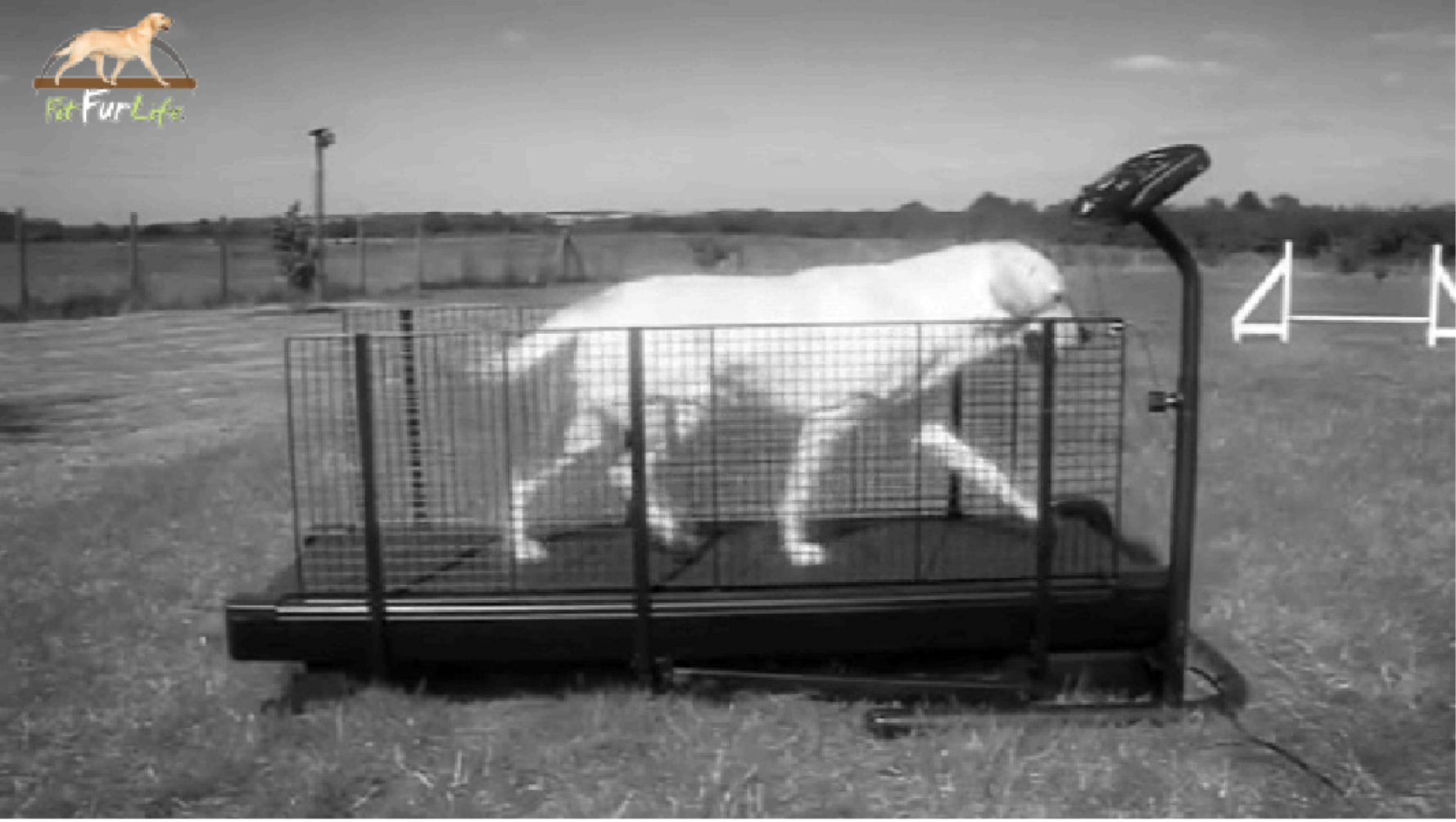}
	\label{fig:suppDog4}
}
\end{center}
\caption{\small{
 Some more examples for the reconstruction achieved for the `dog' dataset. $40\%$ of the test image's pixels (figures \subref{fig:suppDog1} and \subref{fig:suppDog3}) were presented  to the model, which was able to successfully reconstruct them, as can be seen in \subref{fig:suppDog2} and \subref{fig:suppDog4}.
}
}
\label{fig:supplDog}
\end{figure}

\end{document}